\documentclass[10pt,twocolumn,letterpaper]{article}

\usepackage [arxiv]{cvpr} 

\usepackage[accsupp]{axessibility}
\usepackage{url}            
\usepackage{booktabs}       
\usepackage{amsfonts}       
\usepackage{nicefrac}       
\usepackage{microtype}      
\usepackage{multirow, multicol}
\usepackage{colortbl, bm}
\usepackage{graphicx}
\usepackage{amsmath}

\graphicspath{{figures/}}
\newcommand{\comm}[1]{}
\newcommand{\enablepm}[1]{}
\newcommand{\enablespeedup}[1]{}

\renewcommand{\vec}[1]{{\mathbf{#1}}}

\newcommand{\R}{\mathbb{R}}

\newcommand{\fl}{\vec{f}_{\ell}}
\newcommand{\vl}{v_{\ell}}
\newcommand{\ourfacto}{\text{nerfacto}^\dagger}

\definecolor{one}{rgb}{0.0,0.0,0.4}
\definecolor{two}{rgb}{0.0,0.0,0.28}
\definecolor{three}{rgb}{0.0,0.0,0.16}
\definecolor{failure}{rgb}{0.4,0.0,0.0}
\newcommand{\tone}{\cellcolor{blue!42}}
\newcommand{\ttwo}{\cellcolor{blue!28}}
\newcommand{\ttri}{\cellcolor{blue!16}}
\newcommand{\fail}{\cellcolor{red!42}}

\usepackage[dvipsnames]{xcolor}

\definecolor{cvprblue}{rgb}{0.21,0.49,0.74}
\usepackage[pagebackref,breaklinks,colorlinks,citecolor=cvprblue]{hyperref}

\def\paperID{9} 
\def\confName{CVPR}
\def\confYear{2024}

\title{Analyzing the Internals of Neural Radiance Fields}

\author{Lukas Radl \quad {Andreas Kurz} \quad {Michael Steiner} \quad {Markus Steinberger}\\
{\tt\small \{lukas.radl, andreas.kurz,  michael.steiner, steinberger\}@icg.tugraz.at} \\
Graz University of Technology, Austria \\
}

\begin{document}
\twocolumn[{
\maketitle
\begin{center}
    \centering
    \captionsetup{type=figure}
    \includegraphics[width=\textwidth]{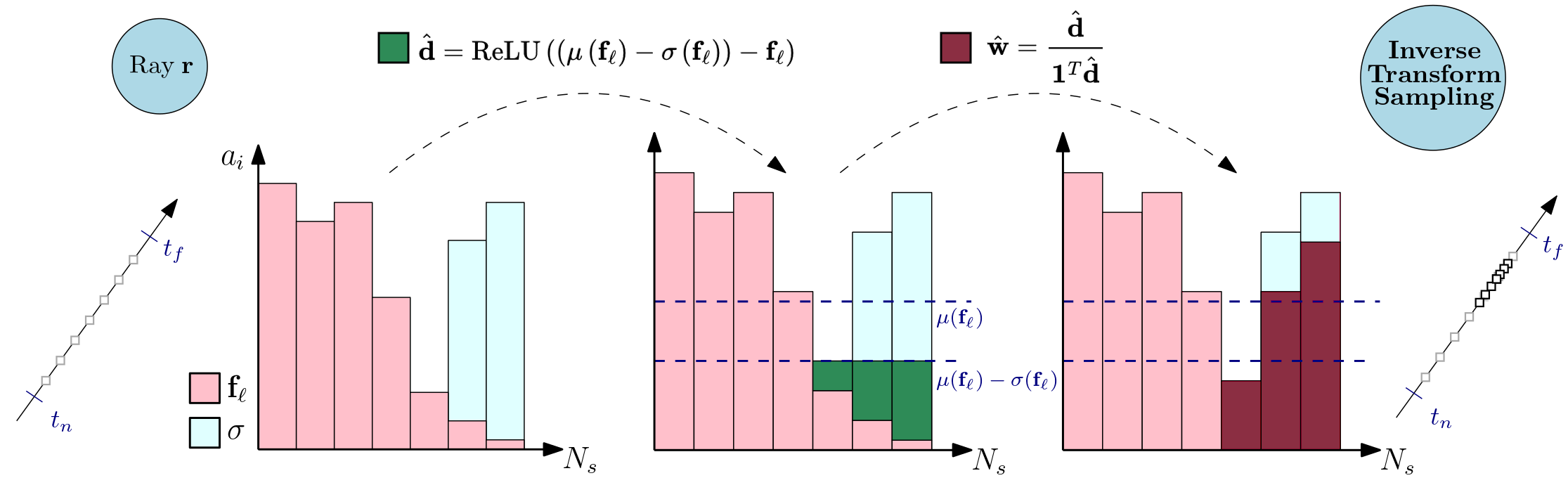}
    \captionof{figure}{
    We show how density estimates derived from intermediate activations can accelerate inference for pre-trained Neural Radiance Fields by effectively reducing the capacity of large MLPs.
    Here, we show a small toy example for a ray with seven, uniformly placed samples between $t_n$ and $t_f$.
    We obtain an activation feature vector $\fl$ for layer $\ell$, apply a function to obtain a density estimate $\hat{\vec{d}}$, leveraging the observation that minima in activation feature space indicate samples with high density $\sigma$. 
    Finally, we perform inverse transform sampling with our weight estimate $\hat{\vec w}$.
    }
    \label{fig:act_informed}
\end{center}%
}]

\begin{abstract}
  Modern Neural Radiance Fields (NeRFs) learn a mapping from position to volumetric density leveraging proposal network samplers. 
  In contrast to the coarse-to-fine sampling approach with two NeRFs, this offers significant potential for acceleration using lower network capacity.
  Given that NeRFs utilize most of their network capacity to estimate radiance, they could store valuable density information in their parameters or their deep features.
  To investigate this proposition, we take one step back and analyze large, trained ReLU-MLPs used in coarse-to-fine sampling.
  Building on our novel activation visualization method, we find that trained NeRFs, Mip-NeRFs and proposal network samplers map samples with high density to local minima along a ray in activation feature space.
  We show how these large MLPs can be accelerated by transforming intermediate activations to a weight estimate, without any modifications to the training protocol or the network architecture.
  With our approach, we can reduce the computational requirements of trained NeRFs by up to $50\%$ with only a slight hit in rendering quality.
  Extensive experimental evaluation on a variety of datasets and architectures demonstrates the effectiveness of our approach. 
  Consequently, our methodology provides valuable insight into the inner workings of NeRFs.
\end{abstract}

\section{Introduction}
\label{sec:intro}
Neural Fields~\cite{XieNeuralFields} parameterize implicit functions over continuous domains using neural networks and have gained popularity due to their ability to represent various types of signals.
To represent arbitrary high-frequency functions, positional encoding~\cite{MildenhallNeRF}, Fourier features~\cite{TancikFourier}, Gaussian activations~\cite{chng2022gaussian, ramasinghe2022beyond}, among other alternatives~\cite{fathony2020multiplicative, saragadam2023wire, sitzmannSiren}, are used to overcome the spectral bias~\cite{rahamanSpectral}. 
Compared to their explicit counterparts, implicit representations are more memory-efficient and can achieve higher quality, but regrettably suffer from slower inference.
In addition, as the signal is encoded in the weights and biases of a neural network, every change to a parameter has a non-local effect, which makes implicit representations hard to modify.

For novel view synthesis problems, the 5D plenoptic function mapping positions and directions to radiance and density can be learned from posed images leveraging a neural field. 
NeRFs~\cite{MildenhallNeRF} have recently demonstrated impressive performance for real-world and synthetic datasets.
However, the na\"ive approach of sampling the scene densely is slow and inefficient due to the prohibitive number of required samples.
A coarse-to-fine sampling strategy is adopted to overcome these limitations, estimating the density distribution along a ray by uniformly sampling a coarse NeRF, which generates a new set of samples for a subsequent fine NeRF using inverse transform sampling. 
Although performance is improved substantially, the number of required samples still prohibits real-time rendering. 
Furthermore, as no ground-truth density information is available, the coarse NeRF is supervised with an MSE reconstruction loss.

Reducing the number of required samples for NeRFs is a fruitful endeavor to lower the computational requirements.
Sampling networks~\cite{arandjelovic2021nerfindetail, Barron2022MipNeRF360, KurzAdaNeRF, NeffDONeRF}, which predict volumetric density along a ray to produce a set of samples for a subsequent {shading} network, prove highly effective for this task. 
The network capacity for a {sampling} network is significantly reduced, as solely predicting volumetric density is view-independent and thus easier to learn.
For supervising these {sampling} networks ground-truth depth data~\cite{NeffDONeRF}, a pre-trained NeRF~\cite{PialaTerminerf}, multi-view stereo~\cite{lin2022efficient} or distillation from a concurrently trained {shading} network~\cite{arandjelovic2021nerfindetail, Barron2022MipNeRF360, KurzAdaNeRF} may be used.
Alternatively, occupancy grids~\cite{MuellerNGP} allow for efficient empty-space skipping by encoding the expected density within a discretized region.


A clear disadvantage of the coarse-to-fine NeRF pipeline is the use of the {coarse} model for density prediction only, even though it is optimized to reconstruct the ground-truth image.
Sampling networks, on the other hand, map sample positions to volumetric density using a network with reduced capacity and can be supervised by the shading network.
Given that sampling networks show that density can be modelled with lower capacity, a subset of a coarse NeRF should be able to model density accurately as well --- particularly since recent work has found that NeRF-MLPs naturally partition themselves into structural and color layers~\cite{wang2023inv}. 

To this end, we analyze the intermediate activations of coarse NeRFs:
We derive a representation from these activations and show how it is tied to the density distribution along a view ray.
Building on this crucial finding, we devise a novel method to extract density estimates given intermediate activations of coarse NeRFs.
By elegantly combining activation analysis and our method to extract densities, our approach can reduce inference times while maintaining high quality at test-time.
We demonstrate that our approach is not only applicable to NeRF and Mip-NeRF~\cite{barron2021mipnerf}, but produces similar results when applied to proposal network samplers as well, with extensive evaluation supporting our claim.
To the best of our knowledge, our approach is the first to use intermediate activations to accelerate inference for coordinate-based MLPs.

To summarize, our contributions are as follows:
\begin{itemize}
    \item We present a novel method for visualizing and analyzing the activations of coordinate-based ReLU-MLPs.
    \item We present an approach for extracting a density estimate for fine re-sampling from activations in early layers of coarse NeRFs, significantly reducing the inference time with little decrease in quality. 
    Consequently, our method provides valuable insight into the inner workings of NeRFs.
    \item We demonstrate the effectiveness of our method for several architectures with real-world and synthetic data.
\end{itemize}

\section{Related Work}
\label{sec:relatedwork}
In the following, we review related work, focusing on real-time rendering and efficient sampling for NeRFs.

\paragraph{NeRFs for Real-time Rendering.}
Instead of the implicit representation of radiance fields used in NeRF~\cite{MildenhallNeRF}, explicit representations with grid-like data structures~\cite{YuPlenoxels, dvgo} or hybrid representation~\cite{TensorRFECCV, Karnewar2022ReLUFields, MuellerNGP, diver, pointnerf} are often used to accelerate inference and training.
Most prominently of the aforementioned, Instant-NGP~\cite{MuellerNGP} replaces the fixed positional encoding used by NeRF with a learnable hash encoding, allowing for the use of much smaller MLPs and thus rapid convergence. 
Alternatively, 3D Gaussian Splatting~\cite{kerbl3Dgaussians} has recently demonstrated state-of-the-art rendering quality in addition to fast training and rendering using a fully explicit representation relying on anisotropic 3D Gaussians.
Other works~\cite{rebainDerf, ReiserKiloNeRF} suggest a divide-and-conquer approach, utilizing many tiny MLPs responsible for a small region instead of a large MLP for the whole scene.
Distillation of a NeRF into a render-friendly format has been explored extensively~\cite{GarbinFastNeRF, hedmanBaking}, but these approaches are memory-intensive and require an additional distillation step after training.
Recent work~\cite{TetraNeRF, DuplexNeRF} has also investigated approximate geometry equipped with neural features and small, global decoders.
Finally, mesh-based view synthesis~\cite{yariv2023bakedsdf, chen2022mobilenerf, reiser2024binary} has recently demonstrated real-time rendering on low-power devices, relying on approximate meshes extracted from radiance fields and the well-known graphics pipeline.

\paragraph{Efficient Sampling.}
Several works investigate a learnable proposal sampler, which can predict volumetric density~\cite{Barron2022MipNeRF360} or sample locations~\cite{arandjelovic2021nerfindetail}.
DONeRF~\cite{NeffDONeRF} learns a density volume as a multi-class classification task to guide sample placement with direct depth supervision.
Building on DONeRF, AdaNeRF~\cite{KurzAdaNeRF} mitigates the requirement of pre-training and depth supervision for the sampling network with a 4-phase training recipe.
EfficientNeRF~\cite{HuEfficient} proposes valid sampling, caching densities of sample positions and exclusively evaluating those with positive density --- this approach shares similarities with our method in that it is motivated by the density distribution of coarse NeRFs.
TermiNeRF~\cite{PialaTerminerf} distills a NeRF into a sampling and shading network. 
For their sampling network, they perform distribution matching, effectively reparameterizing NeRFs intervals to allow for fewer samples. 
Lin~\emph{et al.}~\cite{lin2022efficient} learn a proxy geometry with a cost volume for depth guidance, uniformly sampling a smaller depth range with a lower sample count. 
A better sampling strategy is also attained by unifying implicit surface models and radiance fields~\cite{OechsleUnisurf}.
To enable mesh-based view synthesis, SDFs~\cite{yariv2023bakedsdf} or Binary Opacity Grids~\cite{reiser2024binary} can accurately locate surfaces.
Finally, by restricting volume rendering to a small band around the surface~\cite{adaptiveshells2023}, the number of samples can be reduced significantly.

\section{Preliminaries}
In the following section, we briefly summarize the most important architectures and concepts which build the foundation for our work.

\subsection{NeRF}
\label{sec:3:nerf}
NeRFs~\cite{MildenhallNeRF} learn a function
\begin{equation}
    \Theta_{\text{NeRF}}: \R^5 \rightarrow \R^4,\ (\vec{p}, \vec{d}) \mapsto (\vec{c}, \sigma),
\end{equation}
where $\vec{p} = (x,y,z)$ and $\vec{d} = (\theta, \phi)$ are the sample position and viewing direction, $\vec{c} \in [0,1]^3$ denotes the predicted output color and $\sigma \in \R$ is the volumetric density. 
NeRF uses a ReLU-MLP with 8 layers and $56$ hidden units to predict a feature vector $\vec f \in \R^{256}$ and $\sigma$. 
This feature vector is then concatenated with $\vec d$ and passed through a small ReLU-MLP to produce $\vec c$, which allows view-dependent phenomena to be modelled.
For each pixel, we cast a ray $\vec{r}(t) = \vec{o} + t\vec{d}$ from the origin $\vec o$ in the direction of $\vec d$. 
We evaluate $\Theta_{\text{NeRF}}$ at different sample positions $t \in [t_n, t_f]$ between the near bound $t_n$ and the far bound $t_f$.
The color $\hat{\mathcal{C}}$ of a ray $\vec{r}$ can be estimated using $N_s$ samples with
\begin{equation}
    \hat{\mathcal{C}}(\vec{r}) = \sum_{i=1}^{N_s} T_i (1 - \exp({-\sigma_i \delta_i}))\vec{c}_i,
\end{equation}
where $\delta_i$ and $T_i$ are given as
\begin{equation}
    \delta_i = t_{i+1}-t_i, \quad T_i = \exp{\left(- \sum_{j=1}^{i-1} {\sigma_j \delta_j}\right)}.
\end{equation}
To increase rendering efficiency, two NeRFs are concurrently optimized.
The {coarse} NeRF $\Theta_{\text{coarse}}$ places $N_c$ samples uniformly for each ray $\vec r$, which produces a set of weights $\vec w \in \R^{N_c}$ with
\begin{equation}
     \mathrm{w}_i = { T_i(1 - \exp(-\sigma_i \delta_i))}, \label{eq:3:weights}
\end{equation}
used to express the estimated coarse radiance as a weighted sum
\begin{equation}
    \hat{\mathcal{C}}_c(\vec r) = \sum_{i=1}^{N_c} \mathrm{w}_i \vec c_i.
\end{equation}
The weights $\vec w$ are normalized such that $\vec 1^T \vec w= 1$ to obtain a piecewise-constant PDF along $\vec r$.
This distribution is used to produce a new set of $N_f$ samples with inverse transform sampling.
The final rendered color is produced by $\Theta_{\text{fine}}$ using all $N_c + N_f$ samples.

NeRF applies positional encoding $\gamma(\cdot)$ to $\vec p$ and $\vec d$, allowing the network to model high-frequency details effectively~\cite{TancikFourier}. 
The network is supervised with the MSE loss for both $\Theta_{\text{coarse}}$ and $\Theta_{\text{fine}}$, i.e. $\sum_{\vec r}\|\hat{\mathcal{C}}_c(\vec r) - {\mathcal{C}}(\vec r)\|_2^2 + \|\hat{\mathcal{C}}_f(\vec r) - {\mathcal{C}}(\vec r)\|_2^2$, with $\mathcal{C}(\vec r)$ denoting the ground truth color for $\vec r$ and $\hat{\mathcal{C}}_f(\vec r)$ denoting the estimated {fine} radiance of $\vec r$.


\subsection{Mip-NeRF}
\label{sec:3:mipnerf}

Mip-NeRF~\cite{barron2021mipnerf} builds on NeRF and tackles the problem of aliasing when scene content is observed at various resolutions. 
The standard computer graphics solution of super-sampling is prohibitively expensive when considering NeRFs volume rendering. 
Instead, Mip-NeRF introduces a novel Integrated Positional Encoding (IPE), which is the expected positional encoding of all coordinates within a conical frustum. 
Each conical frustum is approximated with a multivariate Gaussian, and the IPE feature serves as the encoded input for $\Theta_{\text{NeRF}}$.
This modification allows the MLP to reason about the size and shape of the encoded region, efficiently combating aliasing which occurs for ambiguous point samples.

As the network can now learn a multi-scale scene representation, $\Theta_{\text{NeRF}}$ performs both coarse and fine sampling, reducing the model size by $50\%$ --- however, the network is still supervised with an MSE loss.


\subsection{Mip-NeRF 360}
\label{sec:3:mipnerf360}
NeRF and Mip-NeRF require that 3D coordinates exist in a bounded domain for sampling --- an assumption which is violated in unbounded 3D scenes, where objects may exist at any distance from the camera.
Therefore, Mip-NeRF 360~\cite{Barron2022MipNeRF360} proposes a spatial contraction $\text{contract}(\cdot)$, which maps coordinates from $\R^3$ to a ball of radius $2$:
\begin{align}
    \text{contract}(\vec x) = \begin{cases}
        \vec x & \|\vec x \| \leq 1 \\
        \left( 2 - \frac{1}{\|\vec x \|}\right)\left( \frac{\vec x}{\|\vec x \|}\right) & \|\vec x \| > 1 \\
    \end{cases}.
\end{align}
To handle the increased capacity required for unbounded scenes and increase rendering efficiency, Mip-NeRF 360 concurrently optimizes two MLPs, $\Theta_{\text{prop}}$ and $\Theta_{\text{NeRF}}$.
Their proposal network sampler $\Theta_{\text{prop}}$ learns a function
\begin{equation}
        \Theta_{\text{prop}}: \R^3 \rightarrow \R,\ \vec{p} \mapsto \sigma,
\end{equation}
and is optimized to bound the weights $\vec w$ produced by $\Theta_{\text{NeRF}}$. 
Two subsequent rounds of proposal sampling produce two weights $\{\hat{\vec w}_1, \hat{\vec w}_2\}$, which is efficient due to the small size of $\Theta_{\text{prop}}$ with 4 layers and 256 hidden units. 

\section{Activation Analysis}
In this section, we present our method for visualizing and analyzing the intermediate activations for ReLU-MLPs. 
In addition, we show how we can leverage these intermediate activations to accelerate inference for pre-trained NeRFs.

\begin{figure}[t]
    \centering
    \includegraphics[width=\linewidth]{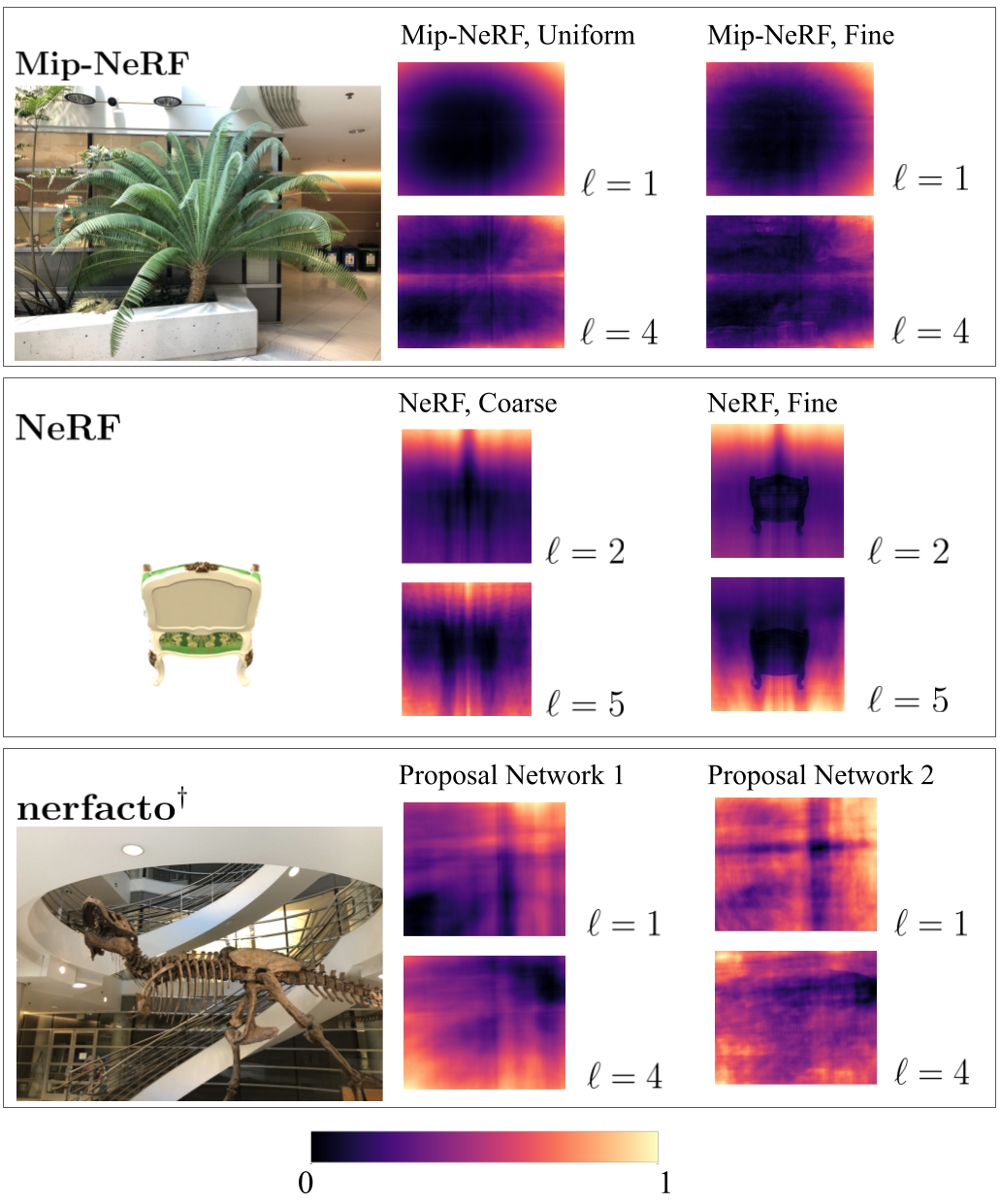}
    \caption{Ground-truth images and their corresponding normalized {coarse} and {fine} activations $\vl$ using the \emph{magma} colormap reveal an interesting relationship between activations and outputs.
    With our visualization approach, we can infer some scene content using only $\vl$.
    For each scene, we visualize activations for different layers $\ell$.}
    \label{fig:activations}
\end{figure}

\subsection{Visualizing and Analyzing Activations}
\label{sec:4:VisualizeDensity}
Benefiting from the inherent spatial structure baked into their building blocks, several works~\cite{SpringenbergAllConv, ZeilerVisualizingCNN} investigate the visualization of intermediate features for Convolutional Neural Networks (CNNs).
These methods typically operate by back-projection from feature space to pixel space, allowing informative insight into the inner workings of these models.
For MLPs, which are mostly used as universal function approximators, practitioners have not been as interested in their activations.
To this end, we propose a novel method for visualizing activations of coordinate-based MLPs.

We analyze the hidden layer activations of NeRF MLPs as follows: For each pixel $(x,y)$, a ray $\vec r$ is generated, positional encoding $\gamma(\cdot)$ is applied and each sample along the ray is passed through $\Theta_{\text{NeRF}}$. In the following, we describe the intermediate activations in terms of a single ray/pixel. After each linear layer, the activation $\vec A^{(\ell)}$ is of size $\R^{N_s \times N_h}$, where $N_s$ and $N_h$ denote the number of samples and the number of hidden units, respectively. 
To obtain a feature representation $\fl \in \R^{N_s}$ for the activation in layer $\ell$, we compute the mean over the dimension $N_h$:
\begin{equation}
     \fl = \frac{1}{N_h}\ \sum_{i=1}^{N_h}\ {\vec{A}^{(\ell)}_{i}}. \label{eq:feature_per_sample}
\end{equation}
Further, we can produce a scalar $\vl \in \R$ for each ray $\vec r$ for visualization purposes with
\begin{equation}
     \vl = \frac{1}{N_s}\ \sum_{i=1}^{N_s}\ \sum_{j=1}^{N_h}\ {\vec{A}^{(\ell)}_{i, j}}. \label{eq:activation}
\end{equation}
We visualize the activations of different layers $\ell$ for some example views using Eqn.~\eqref{eq:activation} in Fig.~\ref{fig:activations}. 
As we can see, intermediate activations exhibit notable structural information.
We note that when recording activations $\vec A^{(\ell)}$, we have the option to apply or not apply the activation function.
Unless stated otherwise, we work with activations after the ReLU has been applied, thus $\vec A^{(\ell)} \in \mathbb{R}_+^{N_s \times N_h}$. 
Although unintuitive due to the information loss caused by the ReLU, early experiments showed that applying the activation function resulted in more consistent histograms for $\fl$.

\begin{figure}[t]
    \centering
    \includegraphics[width=\linewidth]{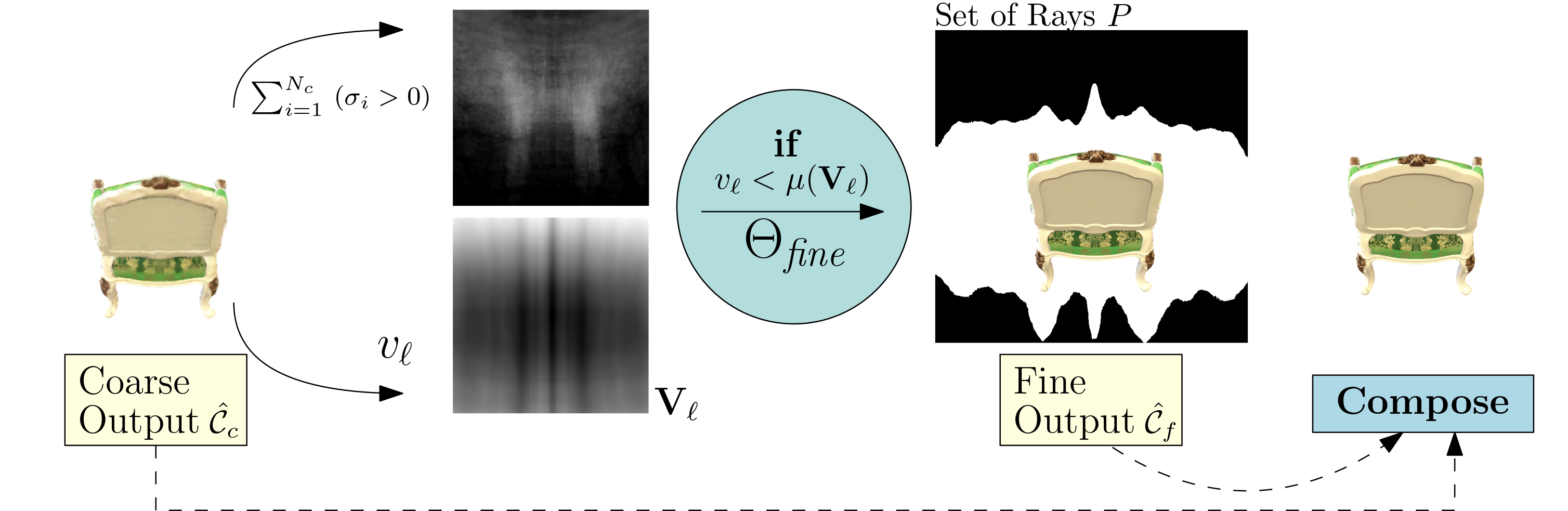}
    \caption{Intermediate activations allow for simple performance improvements.
    We can reduce the inference time of NeRFs in synthetic scenes if we perform the fine pass only if the condition $\vl < \mu(\vec V_\ell)$ is met.}
    \label{fig:fineonly_approach}
\end{figure}
\begin{table}[t]
  \footnotesize
\setlength{\tabcolsep}{4.25pt}
  \centering
    \begin{tabular}{@{}ccccccc@{}}\toprule
    $\ell$  & PSNR $\uparrow$                     & SSIM $\uparrow$         & LPIPS $\downarrow$         & $t\ [s]$              & Speedup       & $\%$ fine Rays \\\midrule
    1       & 29.66 \enablepm{$\pm$ 1.21}          & 0.93 \comm{$\pm$ 0.01} & 0.04  \comm{$\pm$ 0.01} & 34.69 \enablepm{$\pm$ 0.56}    & 40$\%$     & 59$\%$\\
    2       & 30.45 \enablepm{$\pm$ 1.07}          & 0.94  \comm{$\pm$ 0.01} & 0.03  \comm{$\pm$ 0.01} & 34.02 \enablepm{$\pm$ 1.18}    & 43$\%$     & 58$\%$\\
     {3}       & \ttwo {31.21 \enablepm{$\pm$ 1.26}}     & \ttwo 0.95  \comm{$\pm$ 0.01} & \ttwo 0.02  \comm{$\pm$ 0.01} & 33.11 \enablepm{$\pm$ 1.34}    &  {47}$\%$&  55$\%$ \\
    4       &  \ttri 30.71 \enablepm{$\pm$ 1.68}          & \ttri 0.95  \comm{$\pm$ 0.01} & \ttri 0.03  \comm{$\pm$ 0.01} & 34.76 \enablepm{$\pm$ 1.91}    & 43$\%$     & 59${\%}$\\\midrule    
    NeRF    & \tone 32.20 \enablepm{$\pm$ 1.19}          & \tone 0.96  \comm{$\pm$ 0.00} & \tone 0.02  \comm{$\pm$ 0.00} & 48.98 \enablepm{$\pm$ 0.50}    & -         & 100$\%$\\
    \bottomrule
    \end{tabular}
  \caption{Quantitative results for our activation informed coarse-to-fine experiment for the \emph{lego} scene from the Blender dataset~\cite{MildenhallNeRF}. 
  Only performing the fine pass for rays $\vec r$ where the summed activation is smaller than $\tau = \mu(\vec V_\ell)$ yields a significant performance gain with a slight drop in rendering quality.}
  \label{tab:less_fine_evaluations}
\end{table}
\subsection{Reducing Unnecessary Network Evaluations using Activations}
\label{sec:4:synthetic}
As a first step towards speeding up the coarse-to-fine NeRF rendering pipeline, we analyze the densities $\sigma$ predicted by $\Theta_{\text{coarse}}$. 
Take, for example, any of the synthetic scenes from the Blender dataset~\cite{MildenhallNeRF}. 
Along rays $\vec r$ in empty space, we do not expect any sample with significant volumetric density.
We can sum the number of densities along each ray $\vec r$ larger than a small threshold $\tau$ to obtain a mapping $g: \vec r \mapsto \{0, \hdots, N_c\}$. 
If for any of the rays the result is ${0}$, the corresponding pixel likely belongs to the transparent background. 
We see in Fig.~\ref{fig:fineonly_approach} that when the value of $g$ is high, the value of $\vl$ is low, and vice-versa.

For each ray $\vec r$, we record the predicted coarse radiance $\hat{\mathcal{C}}_c$ and the activations $\vl$ from $\Theta_{\text{coarse}}$. 
We choose a threshold $\tau$, which indicates whether we observe scene content along $\vec r$. 
We choose $\tau = \mu(\vec V_\ell)$, the mean activation of the activation image $\vec V_\ell$. 
We obtain a set of rays $P = \{\vec r: \vl < \tau\}$, which we treat as a mask. 
If $\vec r \in P$, we evaluate $\Theta_{\text{fine}}$. 
Otherwise, we treat the coarse radiance $\hat{\mathcal{C}}_c$ as our final output. 
Using this approach, we can significantly reduce the number of evaluations for $\Theta_{\text{fine}}$, which uses $N_c + N_f = 192$ samples, which results in a significant performance improvement.
As we cover most of the object in the scene with our derived mask $P$, the image quality loss is negligible.
We report the results for the \emph{lego} scene from the Blender dataset~\cite{MildenhallNeRF} in Tab.~\ref{tab:less_fine_evaluations} and visualize our approach in Fig.~\ref{fig:fineonly_approach}.
This experiment demonstrates that early intermediate activations already contain notable density information and reveals an interesting connection between density and activations.

\begin{figure}[ht!]
    \centering
    \includegraphics[width=\linewidth]{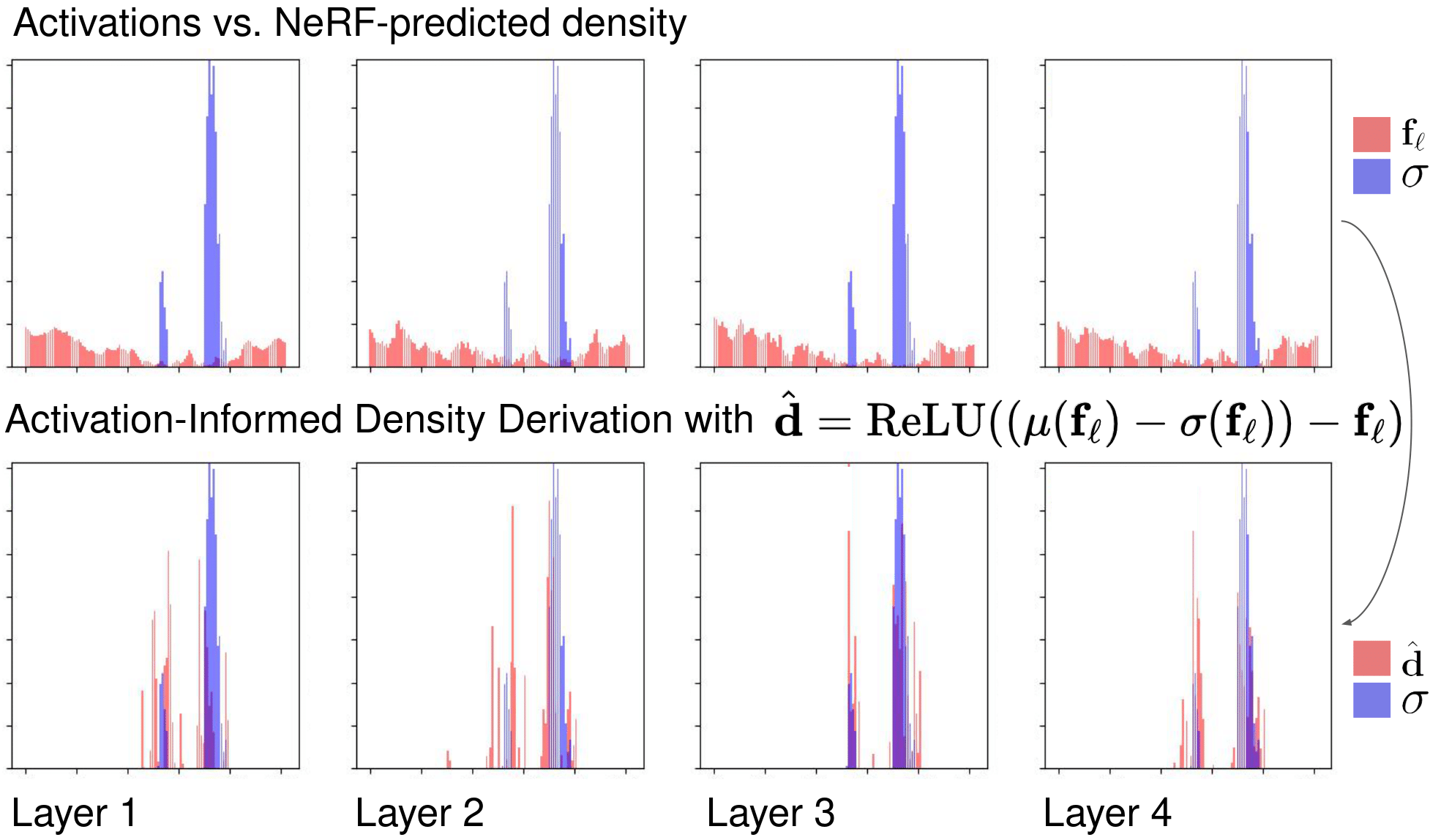}
    \caption{Visualization of our proposed approach for approximate density extraction for a real-world example:
    We visualize activation features $\fl$ and densities $\sigma$ for $128$ uniform samples along an example ray, for a Mip-NeRF trained on the \emph{chair} scene.
    Using Eqn.~\eqref{eq:std}, a plausible density estimate $\hat{\mathbf{d}}$ is extracted from a activation feature $\fl$. }
    \label{fig:real-world-activations}
\end{figure}

\subsection{Approximating Density using Activations}
\label{sec:4:approxdensity}
Clearly, the approach from Sec.~\ref{sec:4:synthetic} heavily relies on the transparent background present in synthetic scenes.
Real-world captures, on the other hand, often include highly detailed backgrounds.
Therefore, we go beyond this basic approach by analyzing the activation features $\fl$ along a ray $\vec r$. 
We accomplish this by examining histograms of the predicted density $\sigma$ and the activation features $\fl$, as can be seen in Fig.~\ref{fig:real-world-activations}.
We notice a trend throughout multiple scenes: Activation feature space minima indicate samples with significant density. 
If we can find a function to transform $\fl$ to a reasonable density estimate $\hat{\vec d}$, we can effectively replace the expensive evaluation for $\Theta_{\text{coarse}}$ with a pass through a smaller MLP.
Ideally, this function should transform $\fl$ to a vector of mostly zeros, except for locations where $\sigma$ is significant.
To this end, we propose 3 different functions $f_i$ to map the intermediate activations directly to estimated densities $\hat{\vec d}$, which were handcrafted based on the observed relationship between activations and density:
\begin{align}
    \comm{\hat{\vec d} =} f_1\left( \fl \right) &= \text{ReLU} \left( \left(\mu(\fl) - \sigma(\fl)\right) - \fl\right), \label{eq:std}\\
    \comm{\hat{\vec d} =} f_2\left( \fl \right)  &= \text{ReLU} \left( \left(\mu(\fl) - \frac{\sigma(\fl)}{2}\right) - \fl\right), \label{eq:stdhalf}\\
    \comm{\hat{\vec d} =} f_3\left( \fl \right)  &= \text{ReLU} \left( \left(\mu(\fl) - \frac{\sigma(\fl)}{2}\right) - \fl\right)^2, \label{eq:stdhalfsq}
\end{align}
where $\fl$ denotes the intermediate activation feature for layer $\ell$ and $\mu(\cdot),\ \sigma(\cdot)$ denote the mean and standard deviation along the last dimension (i.e. along the $N_s$ samples of the ray $\vec r$). 
We visualize our approach for a toy example in Fig.~\ref{fig:act_informed}.

Using Eqn.~\eqref{eq:std} results in fewer intervals with density along the ray, but is prone to failures if the distribution along a ray exhibits high standard deviation.
On the other hand, Eqns.~\eqref{eq:stdhalf}~and~\eqref{eq:stdhalfsq} extract density estimates more conservatively but might lead to insufficient detail, as the interval between fine samples grows larger.

Given a well-behaved distribution, we can estimate densities sufficiently close to $\sigma$ such that we get good reconstruction quality when converting $\hat{\vec d}$ to a weight estimate $\hat{\vec w}$. 
Further, we can significantly reduce the required inference time by using only a small subset of the large $\Theta_{\text{coarse}}$ MLP. 
Clearly, the potential speedup is dependent on the number of samples $N_s$, which is large for NeRFs.

\paragraph{Our Proposed Method for Weight Estimation.}
First, we extract the activation features $\fl$ and apply a function $f_i$ to transform this to a density estimate $\hat{\vec d}$ for each ray $\vec r$. 
To obtain our estimated weights $\hat{\vec w}$, which should conform to a piecewise-constant PDF along the ray $\vec r$, we use $\hat{w}_i = \hat{d}_i / \sum_{j=1}^{N_s}\hat{d}_j $ and subsequently perform inverse transform sampling, as done in NeRF~\cite{MildenhallNeRF} (c.f. Fig.~\ref{fig:act_informed}).

\section{Experiments}
In the following, we apply our presented method to NeRF~\cite{MildenhallNeRF} and Mip-NeRF~\cite{barron2021mipnerf}.
Additionally, we apply our method to proposal network samplers, utilizing a nerfacto model~\cite{tancik2023nerfstudio} with a Mip-NeRF 360~\cite{Barron2022MipNeRF360} proposal sampler, which we dub $\ourfacto$.
We only apply our density extraction method to the first 3 layers:
Recent work by Wang~\emph{et al.}~\cite{wang2023inv} has found that NeRF MLPs naturally store structural information within the first MLP layers.
In addition, when applying our method to earlier layers, we can accelerate inference more substantially.

\paragraph{Datasets.}
For our evaluation, we use four diverse datasets, which are well-established in the novel view synthesis literature:
The \textbf{Blender} dataset~\cite{MildenhallNeRF} consists of $360^\circ$ captures of synthetic objects with complex geometry, realistic non-Lambertian effects and transparent backgrounds in a bounded domain.
\textbf{LLFF}~\cite{mildenhall2019llff} is a dataset of forward-facing, real-world scenes in high resolution.
The \textbf{Mip-NeRF 360} dataset~\cite{Barron2022MipNeRF360} contains $360^\circ$ captures of real-world, unbounded scenes and is the most challenging benchmark in our evaluation.
Finally, we also use 2 large-scale, bounded indoor scenes from \textbf{Deep Blending}~\cite{DeepBlending2018} (\emph{playroom} \& \emph{drjohnson}).
For both LLFF and Mip-NeRF 360, we use the $4\times$ downsampled images.
We report per-scene results in the supplementary material.

\paragraph{Implementation Details.}
We use the NeRF and Mip-NeRF implementations from Nerfstudio~\cite{tancik2023nerfstudio} and use the hyperparameter settings as reported in the respective works~\cite{barron2021mipnerf, MildenhallNeRF}.
To facilitate sampling in NDC, as NeRF does for LLFF~\cite{mildenhall2019llff}, we implement a custom LLFF dataloader adapted from the nerf-pytorch codebase~\cite{lin2020nerfpytorch}.

For our experiments with proposal network samplers, we adapt the nerfacto model from Nerfstudio~\cite{tancik2023nerfstudio}. 
This architecture unifies recent NeRF-related research~\cite{Barron2022MipNeRF360, martinbrualla2020nerfw, MuellerNGP, verbin2022refnerf, wang2021nerfmm} into a single, universally-applicable method.
For our $\ourfacto$ model, we replace the \mbox{\textit{HashMLPDensityField}} with a 4-layer ReLU-MLP with 256 hidden units, following Mip-NeRF 360~\cite{Barron2022MipNeRF360}, and use positional encoding~\cite{MildenhallNeRF}.
Note that this configuration is slower than nerfacto due to the large MLP, however, it enables experiments with our proposed method for proposal network samplers.

\subsection{Speeding up NeRF and Mip-NeRF}
\begin{table}[ht!]
  \scriptsize
  \setlength{\tabcolsep}{2.8pt}
    \begin{tabular}{@{}cccccccccc@{}}\toprule
    \multicolumn{10}{c}{\textbf{NeRF}} \\\midrule
    \multirow{2}{*}{$\ell$}&\multirow{2}{*}{$f$}&\multicolumn{4}{c}{{Blender}, $800 \times 800$} & \multicolumn{4}{c}{{LLFF}, $1008 \times 756$}\\
    \cmidrule(lr){3-6}\cmidrule(lr){7-10}
     & & PSNR                      & SSIM          & LPIPS          \enablespeedup{& $t\ [\text{s}]$ }             & Speedup & PSNR                      & SSIM          & LPIPS          \enablespeedup{& $t\ [\text{s}]$ }           & Speedup    \\\midrule
    \multirow{3}{*}{1} & $f_1$      & 27.48 \enablepm{$\pm$ 1.19} & 0.92  & 0.09 \enablespeedup{& 25.41} & \multirow{3}{*}{24$\%$}
                                    & 25.11  \enablepm{$\pm$ 0.62} & 0.75  & 0.23 \enablespeedup{& 27.94}  & \multirow{3}{*}{28$\%$} \\
    &$f_2$      & 28.69  \enablepm{$\pm$ 1.10} & 0.92 & 0.08 \enablespeedup{& 25.41} & 
                & \ttwo 25.40  \enablepm{$\pm$ 0.66} & \ttwo 0.76 & \ttwo 0.22 \enablespeedup{& 27.93} &  \\
    &$f_3$      & 28.63  \enablepm{$\pm$ 1.08} & 0.92 & 0.08 \enablespeedup{& 25.42} & 
                & \ttri 25.34  \enablepm{$\pm$ 0.65} & \ttri 0.75 & \ttri 0.23 \enablespeedup{& 28.02}  &  \\\midrule
    \multirow{3}{*}{2}  & $f_1$     & 27.99  \enablepm{$\pm$ 1.73} & 0.92 & 0.08 \enablespeedup{& 26.21}  & \multirow{3}{*}{20$\%$}
                                    & 24.72  \enablepm{$\pm$ 0.57} & 0.73 & 0.26 \enablespeedup{& 28.85}  & \multirow{3}{*}{24$\%$} \\
    &$f_2$      & \ttri29.05 \enablepm{$\pm$ 1.14} & \ttri0.93 & \ttri0.07 \enablespeedup{& 26.22}  & 
                & 25.12 \enablepm{$\pm$ 0.62} & 0.74 & 0.24 \enablespeedup{& 28.89}  &  \\
    &$f_3$      & \ttwo 29.44 \enablepm{$\pm$ 1.16} & \ttwo0.93 & \ttwo0.07 \enablespeedup{& 26.22}  & 
                & 25.01 \enablepm{$\pm$ 0.61} & 0.74 & 0.25 \enablespeedup{& 28.87}  &  \\\midrule
    \multirow{3}{*}{3} & $f_1$      & \fail 24.17 \enablepm{$\pm$ 2.45} & 0.88 & 0.12 \enablespeedup{& 26.67}  & \multirow{3}{*}{18$\%$}
                                    & 24.21 \enablepm{$\pm$ 0.49} & 0.71 & 0.28 \enablespeedup{& 29.76} & \multirow{3}{*}{20$\%$} \\
    &$f_2$      & 26.23 \enablepm{$\pm$ 1.45} & 0.90 & 0.11 \enablespeedup{& 26.70}  & 
                & 24.99 \enablepm{$\pm$ 0.59} & 0.74 & 0.25 \enablespeedup{& 29.86}  & \\
    &$f_3$      & 25.07 \enablepm{$\pm$ 1.75} & 0.89 & 0.11 \enablespeedup{& 26.70}  & 
                & 24.87 \enablepm{$\pm$ 0.58} & 0.73 & 0.26 \enablespeedup{& 29.80}  &  \\\midrule
    \multicolumn{2}{c}{NeRF}        & \tone 29.87 \enablepm{$\pm$ 1.25} & \tone 0.94 & \tone 0.06 \enablespeedup{& 31.48} & -
                                    & \tone 25.89 \enablepm{$\pm$ 0.73} & \tone 0.78 & \tone 0.18 \enablespeedup{& 35.80} & - \\
    \midrule
    \multicolumn{10}{c}{\textbf{Mip-NeRF}} \\\midrule
    \multirow{3}{*}{1} & $f_1$      & 27.54 \enablepm{$\pm$ 1.45} & 0.92 & 0.09 \enablespeedup{& 21.97} & \multirow{3}{*}{58$\%$}
                                    & 25.07 \enablepm{$\pm$ 0.66} & 0.74 & 0.24 \enablespeedup{& 29.12}  & \multirow{3}{*}{50$\%$} \\
    &$f_2$      & 28.75 \enablepm{$\pm$ 1.15} & 0.92 & 0.08 \enablespeedup{& 21.99} &
                & \ttri 25.45 \enablepm{$\pm$ 0.70} & \ttwo 0.76 & \ttwo 0.22  \enablespeedup{& 29.08}&  \\
    &$f_3$      & 28.72 \enablepm{$\pm$ 1.15} & 0.93 & 0.08 \enablespeedup{& 21.98} &
                & 25.35 \enablepm{$\pm$ 0.69} & 0.75 & 0.22 \enablespeedup{& 28.93} &  \\\midrule
    \multirow{3}{*}{2} & $f_1$      & 26.67 \enablepm{$\pm$ 1.82} & 0.91 & 0.09 \enablespeedup{& 23.38} & \multirow{3}{*}{49$\%$}
                                    & 25.02 \enablepm{$\pm$ 0.68} & 0.74 & 0.25 \enablespeedup{& 31.09} & \multirow{3}{*}{40$\%$} \\
    &$f_2$      & \ttri29.04 \enablepm{$\pm$ 1.17} & \ttri0.93 & \ttri0.07 \enablespeedup{& 23.40} & 
                & 25.44 \enablepm{$\pm$ 0.70} & 0.75 & 0.22 \enablespeedup{& 31.28}  &  \\
    &$f_3$      & \ttwo29.35 \enablepm{$\pm$ 1.19} & \ttwo0.93 & \ttwo0.07 \enablespeedup{& 23.40} & 
                & 25.35 \enablepm{$\pm$ 0.69} & 0.75 & 0.23 \enablespeedup{& 31.51} &  \\\midrule
    \multirow{3}{*}{3} & $f_1$      & \fail 23.47 \enablepm{$\pm$ 4.45} & 0.90 & 0.10 \enablespeedup{& 25.05} & \multirow{3}{*}{39$\%$}
                                    & 24.21 \enablepm{$\pm$ 0.49} & 0.71 & 0.28 \enablespeedup{& 29.76} & \multirow{3}{*}{35$\%$} \\
    &$f_2$      & 27.47 \enablepm{$\pm$ 1.54} & 0.92 & 0.09 \enablespeedup{& 25.06} & 
                & \ttwo 25.49 \enablepm{$\pm$ 0.70} & \ttri 0.76 & \ttri 0.22 \enablespeedup{& 32.14}  &  \\
    &$f_3$      & 26.45 \enablepm{$\pm$ 1.96} & 0.91 & 0.09 \enablespeedup{& 25.07} &  
                & 25.42 \enablepm{$\pm$ 0.69} & 0.75 & 0.22 \enablespeedup{& 32.26}  &  \\\midrule
    \multicolumn{2}{c}{Mip-NeRF}    & \tone 29.69 \enablepm{$\pm$ 1.24} & \tone0.94 & \tone0.06 \enablespeedup{& 34.79} & -
                                    & \tone 25.99 \enablepm{$\pm$ 0.74} & \tone 0.78 & \tone 0.18 \enablespeedup{& 43.82}  & - \\
    \midrule
    \multicolumn{10}{c}{{$\textbf{nerfacto}^\dagger$}} \\\midrule
    \multirow{3}{*}{1} & $f_1$      & 25.67 \enablepm{$\pm$ 0.30} & 0.91 & 0.10 \enablespeedup{& 9.90}  & \multirow{3}{*}{83$\%$}          
                                    & 24.46 \enablepm{$\pm$ 0.79} & 0.81 & 0.15 \enablespeedup{& 11.44} & \multirow{3}{*}{79$\%$} \\
    &$f_2$      & 26.20 \enablepm{$\pm$ 0.33} & 0.91 & 0.09 \enablespeedup{& 9.93}  &                            
                & \ttri 24.69 \enablepm{$\pm$ 0.75} & \ttri 0.81& \ttri 0.14 \enablespeedup{& 11.44} &  \\
    &$f_3$      & 26.17 \enablepm{$\pm$ 0.34} & 0.91 & 0.09 \enablespeedup{& 9.92}  &                              
                & 24.13 \enablepm{$\pm$ 0.70} & 0.80 & 0.16 \enablespeedup{& 11.46} &  \\\midrule
    \multirow{3}{*}{2} & $f_1$      & 26.24 \enablepm{$\pm$ 0.26} & 0.91 & 0.09 \enablespeedup{& 14.03}  & \multirow{3}{*}{29$\%$}    
                                    & 24.21 \enablepm{$\pm$ 0.86} & 0.80 & 0.16 \enablespeedup{& 15.90} & \multirow{3}{*}{29$\%$} \\
    &$f_2$      & \ttwo 26.56 \enablepm{$\pm$ 0.24} & \ttwo 0.91 & \ttwo 0.09 \enablespeedup{& 14.04}  &           
                & \ttwo 24.83 \enablepm{$\pm$ 0.83} & \ttwo 0.81 & \ttwo 0.14 \enablespeedup{& 15.90} &  \\
    &$f_3$      & \ttri 26.31 \enablepm{$\pm$ 0.22} & \ttri 0.91 & \ttri 0.09 \enablespeedup{& 14.04} &          
                & 23.51 \enablepm{$\pm$ 0.64} & 0.78 & 0.18 \enablespeedup{& 15.90} &  \\\midrule
    \multirow{3}{*}{3} & $f_1$      & 26.16 \enablepm{$\pm$ 0.33} & 0.91 & 0.09 \enablespeedup{& 18.15}  & \multirow{3}{*}{-}      
                                    & 23.21 \enablepm{$\pm$ 0.58} & 0.78 & 0.19 \enablespeedup{& 20.39} & \multirow{3}{*}{-} \\
    &$f_2$      & 25.58 \enablepm{$\pm$ 0.27} & 0.91 & 0.10 \enablespeedup{& 18.18}  &                           
                & 23.78 \enablepm{$\pm$ 0.79} & 0.79 & 0.17 \enablespeedup{& 20.39} &  \\
    &$f_3$      & 24.79 \enablepm{$\pm$ 0.27} & 0.90 & 0.10 \enablespeedup{& 18.17}  &                            
                & 23.04 \enablepm{$\pm$ 0.66} & 0.77 & 0.22 \enablespeedup{& 20.43} &  \\\midrule
    \multicolumn{2}{c}{$\text{nerfacto}^\dagger$}    & \tone 27.61 \enablepm{$\pm$ 0.31} & \tone 0.93 & \tone 0.07 \enablespeedup{& 18.15}  & -
                                    & \tone 25.35 \enablepm{$\pm$ 0.82} & \tone 0.83 & \tone 0.12 \enablespeedup{& 20.54} & - \\
    \bottomrule
    \end{tabular}
  \caption{Quantitative evaluation of our approach for {NeRF}, {Mip-NeRF} and {$\text{nerfacto}^\dagger$}. 
  For each dataset, we report PSNR, SSIM~\cite{WangSSIM} and LPIPS~\cite{ZhangLPIPS}. 
  We highlight {\colorbox{blue!42}{best}}, {\colorbox{blue!28}{second-best}} and {\colorbox{blue!16}{third-best}} and point out {\colorbox{red!42}{failures}} of our method.}
  \label{tab:results}
\end{table}

We evaluate our approach for NeRF and Mip-NeRF using the Blender dataset~\cite{MildenhallNeRF} and the LLFF dataset~\cite{mildenhall2019llff}.
For the NeRF experiment, we could either record the activations from $\Theta_{\text{coarse}}$ or $\Theta_{\text{fine}}$. 
However, only $\Theta_{\text{coarse}}$ is trained for uniform samples, hence we use the coarse MLP, which also performed better in our experiments.

We report quantitative results in Tab.~\ref{tab:results} and show our visual results in Fig.~\ref{fig:qualitative}. 
For NeRF, the performance gain is not as significant as for Mip-NeRF due to the different number of samples $N_c, N_f$: 
For Mip-NeRF, $N_c = N_f = 128$, contrary to $N_c=64, N_f = 128+64$ for NeRF.
Hence, our approach performs favorably for Mip-NeRF due to its more expensive coarse network evaluation.
We find that Eqn.~\eqref{eq:stdhalf} for $\ell=2$ performs best in our experiments.
All run-time results for NeRF and Mip-NeRF were obtained on an NVIDIA Quadro RTX 8000.

\subsection{Speeding up Proposal Networks}
\begin{figure}[ht!]
    \centering
    \includegraphics[width=\linewidth]{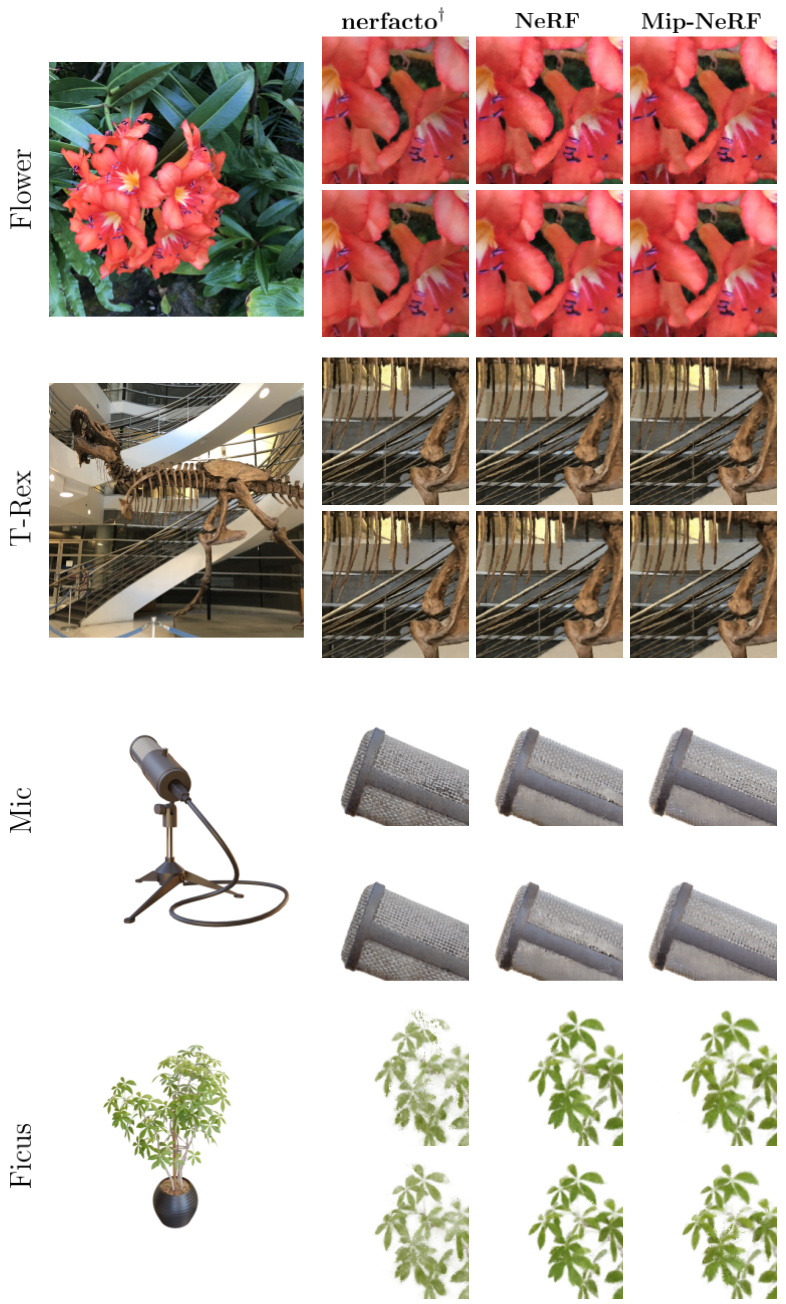}
    \caption{Qualitative results for our approach (top row) compared to baseline methods (bottom row) for synthetic and real-world scenes.
    As can be seen in the zoomed-in views, our best renderings are virtually indistinguishable from the baseline in most configurations.
    }
    \label{fig:qualitative}
\end{figure}

First, we evaluate our approach for $\ourfacto$ using the Blender dataset~\cite{MildenhallNeRF} and the LLFF dataset~\cite{mildenhall2019llff}.
For our experiments, we approximate the first iteration of proposal sampling: 
As the proposal sampler iterations use $\{256, 96\}$ samples, respectively, approximating the first round yields more performance improvements\footnote{Please note that $\ourfacto$ is significantly slower than nerfacto due to the use of an MLP.}.
In addition, our functions $f_i$ are biased towards uniform or piecewise-linear sample placement, and lead to unfavorable results when interval sizes vary significantly.

We report quantitative results in Tab.~\ref{tab:results} and show our visual results in Fig.~\ref{fig:qualitative}. 
For the $\text{nerfacto}^\dagger$ model, we observe a huge performance gain for the first layer, which declines sharply --- this is due to the large number of samples for the first proposal sampler iteration and the small shading network. 
We find that Eqn.~\eqref{eq:stdhalf} applied in layer $\ell=2$ performed best when considering standard image metrics. 
The reported run-time results for $\ourfacto$ were obtained on an NVIDIA RTX 2070 Super.

\begin{figure}
\centering
\rotatebox{90}{
 \begin{minipage}[c]{0.1\linewidth}
    \centering
    \footnotesize
    \text{Ours}
\end{minipage}}
    \includegraphics[width=.30\linewidth]{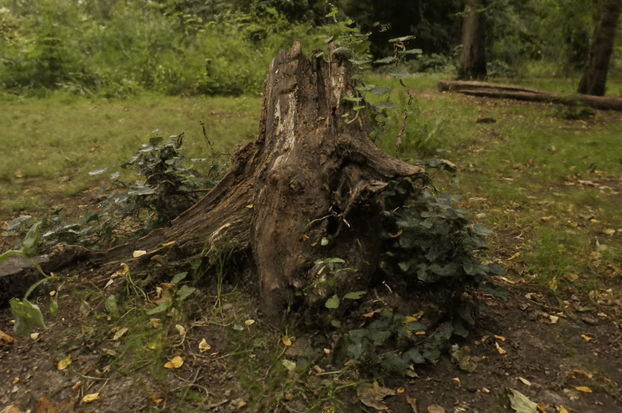}
    \includegraphics[width=.30\linewidth]{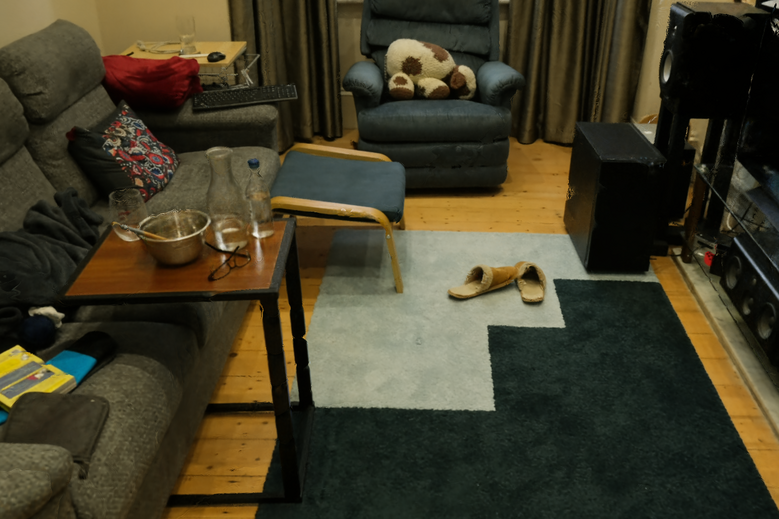}
    \includegraphics[width=.30\linewidth]{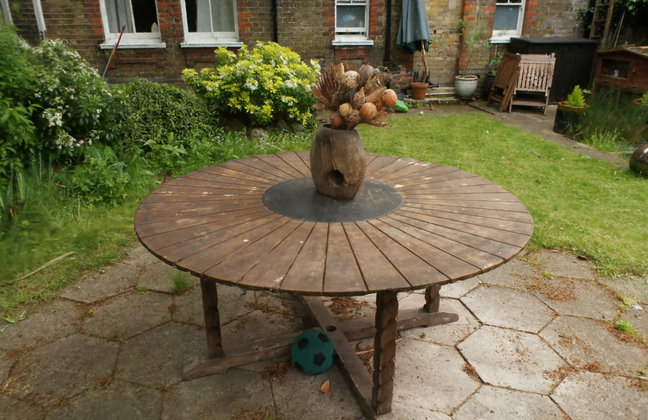} \\
\rotatebox{90}{
 \begin{minipage}[c]{0.1\linewidth}
    \centering
    \footnotesize
    \text{Baseline}
\end{minipage}}
    \includegraphics[width=.30\linewidth]{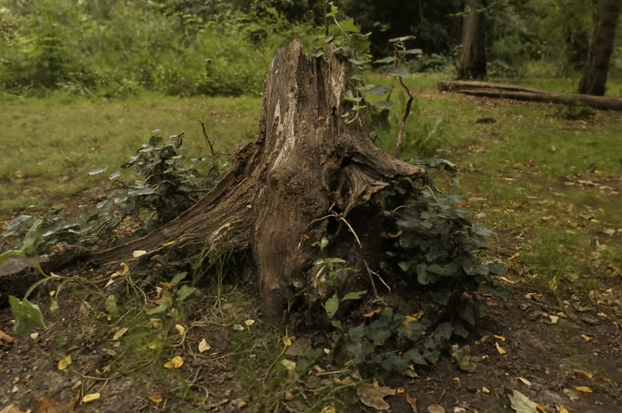}
    \includegraphics[width=.30\linewidth]{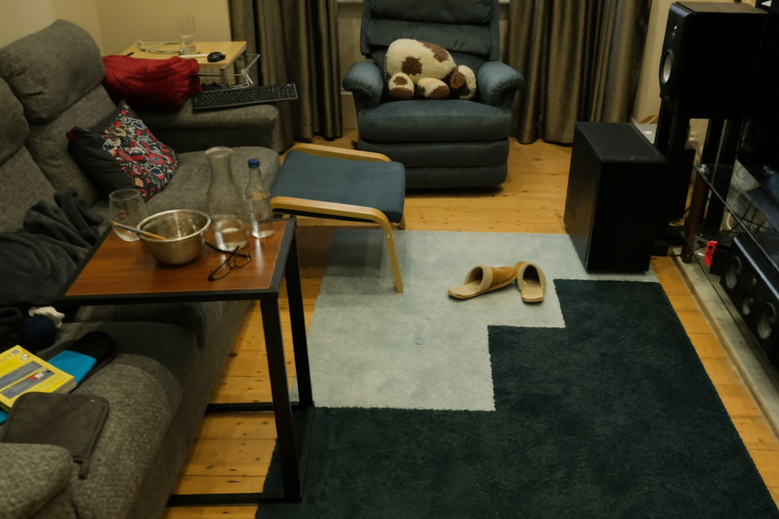}
    \includegraphics[width=.30\linewidth]{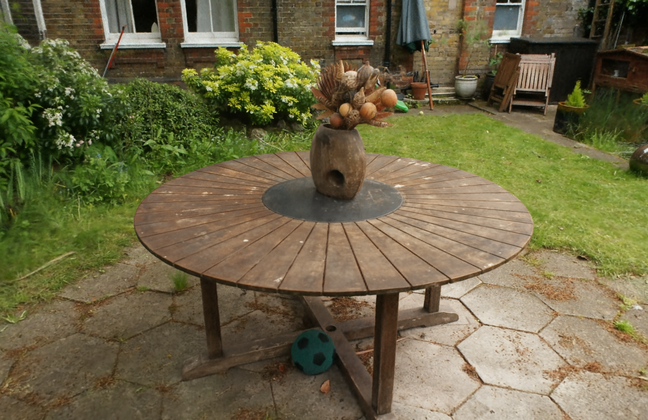} \\
\rotatebox{90}{
 \begin{minipage}[c]{0.1\linewidth}
    \centering
    \footnotesize
    \text{Ours}
\end{minipage}}
    \includegraphics[width=.30\linewidth]{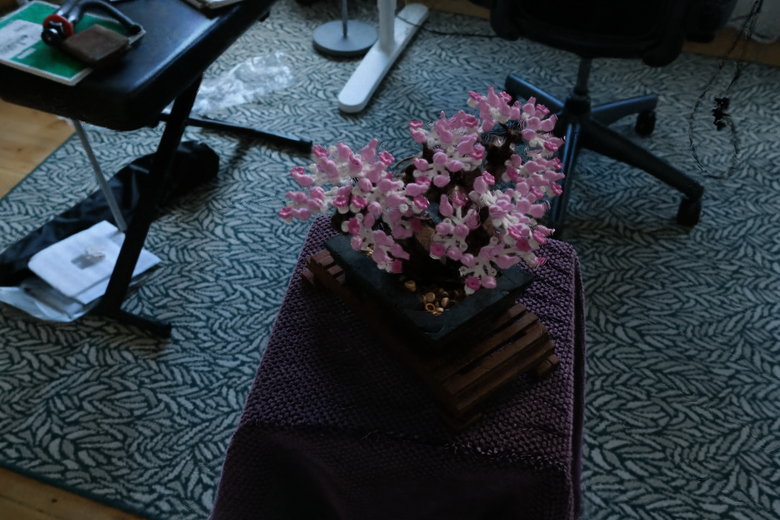}
    \includegraphics[width=.30\linewidth]{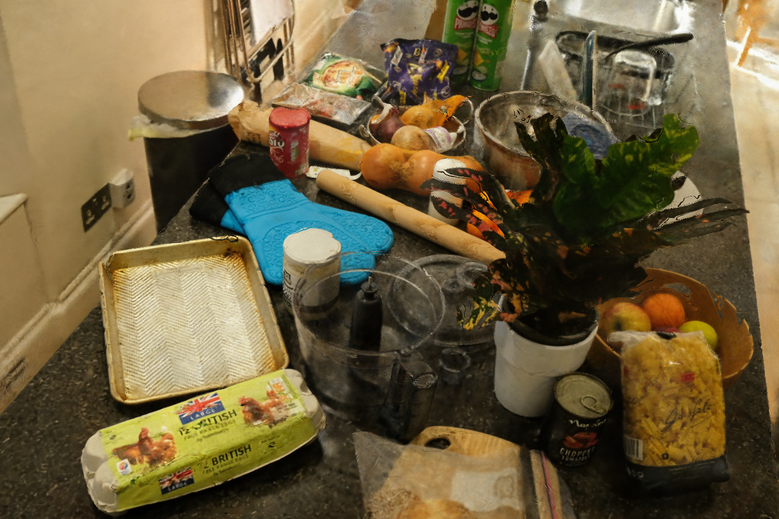}
    \includegraphics[width=.30\linewidth]{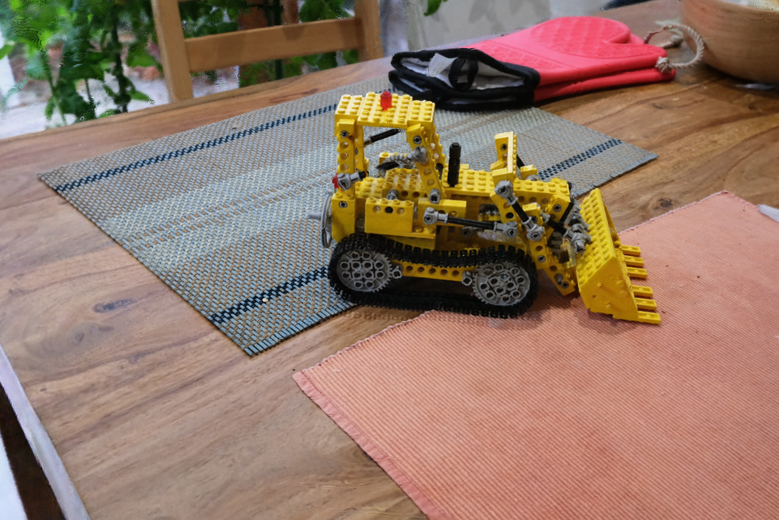} \\
\rotatebox{90}{
 \begin{minipage}[c]{0.1\linewidth}
    \centering
    \footnotesize
    \text{Baseline}
\end{minipage}}
    \includegraphics[width=.30\linewidth]{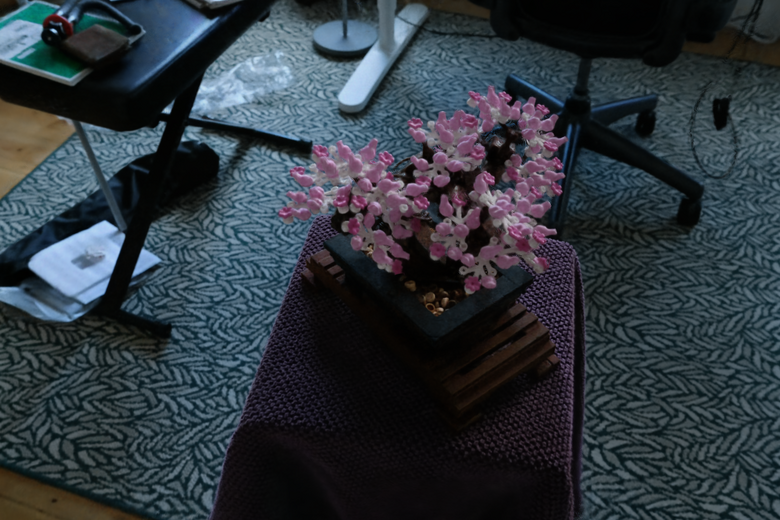}
    \includegraphics[width=.30\linewidth]{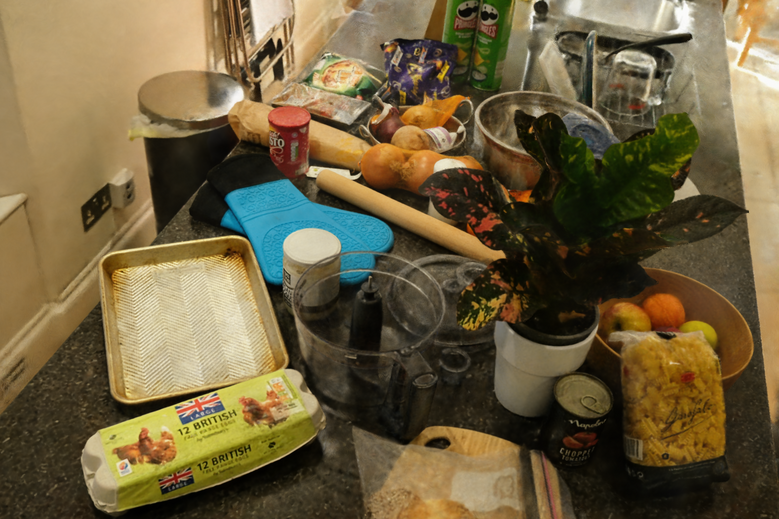}
    \includegraphics[width=.30\linewidth]{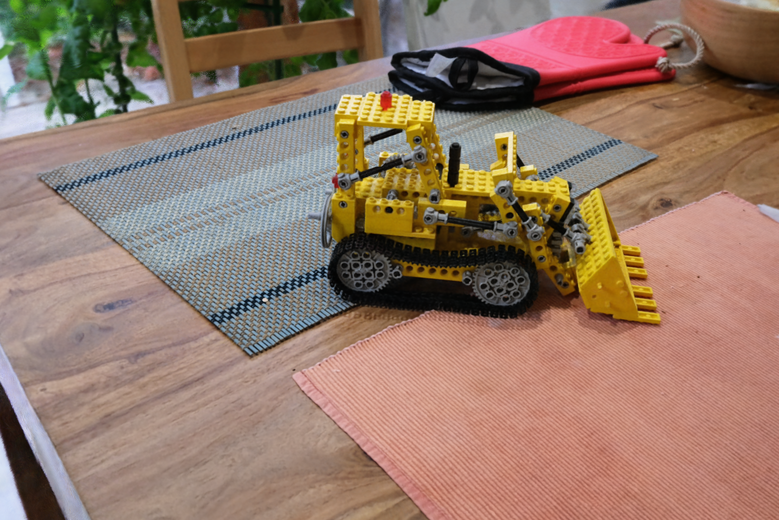} 
    \caption{Example renderings from our our method compared to $\ourfacto$ (Baseline) for the Mip-NeRF 360 dataset~\cite{Barron2022MipNeRF360}.
    In most configurations, our approach is able to retain high visual fidelity on this challenging benchmark.}
    \label{fig:visual-mip}
\end{figure}

Further, we evaluate our method on unbounded, real-world scenes, leveraging the Mip-NeRF 360~\cite{Barron2022MipNeRF360} and Deep Blending~\cite{DeepBlending2018} datasets.
We adapt $\ourfacto$ in the following ways:
We use $7 \times 10^{5}$ training iterations, disable the appearance embedding, camera optimizer, distortion loss and the collider.
We present our results in Tab.~\ref{tab:results_mip360} and show example outputs in Fig.~\ref{fig:visual-mip}.
Compared to the results presented for the LLFF dataset~\cite{mildenhall2019llff}, our method performs less favorably in this scenario.
We attribute this to the large sampling domain in these datasets, which requires more precise density estimates. 
Performance-wise, we observe the same trends as for the previous experiments.
We used an NVIDIA RTX 4090 for these two datasets.

\begin{table}[t]
  \scriptsize
  \setlength{\tabcolsep}{2.8pt}
  \centering
    \begin{tabular}{@{}cccccccccc@{}}
    \toprule
    && \multicolumn{4}{c}{Mip-NeRF 360} & \multicolumn{4}{c}{Deep Blending} \\
    \cmidrule(lr){3-6} \cmidrule(lr){7-10} 
     && PSNR & SSIM & LPIPS  & Speedup & PSNR & SSIM & LPIPS  & Speedup\\
    \midrule
    \multirow{3}{*}{1} &$f_1$ & 22.80 & 0.687 & \ttri 0.282 &   & \ttri 28.01 & \ttri 0.858 & \ttwo 0.237 &  \\
    &$f_2$ & \ttwo 22.92 & \ttwo 0.691 & \ttwo 0.279 &  76 \% & \ttwo 28.01 & \ttwo 0.859 & \ttri 0.237 &  72 \% \\
    & $f_3$ & \ttri 22.80 & \ttri 0.687 & 0.283 &   & 27.98 & 0.858 & 0.239 &  \\
    \midrule
    \multirow{3}{*}{2} &$f_1$ & 22.71 & 0.683 & 0.289 &  & 27.97 & 0.858 & 0.239 &\\
    &$f_2$ & 22.66 & 0.681 & 0.292 & 28 \% & 27.95 & 0.856 & 0.243 & 27 \% \\
    &$f_3$ & 22.67 & 0.680 & 0.297 & & 27.94 & 0.856 & 0.242 &  \\
    \midrule
    \multirow{3}{*}{3} &$f_1$ & 22.44 & 0.675 & 0.305 &  & 27.77 & 0.849 & 0.256 &  \\
    &$f_2$ & 22.21 & 0.669 & 0.309 & - & 27.48 & 0.844 & 0.264 & - \\
     &$f_3$ & 22.39 & 0.669 & 0.316 & & 27.78 & 0.847 & 0.262 & \\\midrule
    \multicolumn{2}{c}{$\text{nerfacto}^\dagger$} & \tone 24.20 & \tone 0.732 & \tone 0.237 & - & \tone 29.26 & \tone 0.880 & \tone 0.207 & - \\
    \bottomrule
    \end{tabular}
  \caption{Quantitative evaluation of our approach for {$\text{nerfacto}^\dagger$} for Mip-NeRF 360~\cite{Barron2022MipNeRF360} and Deep Blending~\cite{DeepBlending2018}. 
  For each dataset, we report PSNR, SSIM~\cite{WangSSIM} and LPIPS~\cite{ZhangLPIPS}.
  We highlight {\colorbox{blue!42}{best}}, {\colorbox{blue!28}{second-best}} and {\colorbox{blue!16}{third-best}} for each image metric.}
  \label{tab:results_mip360}
\end{table}

\subsection{Ablation studies}
\label{sec:ablation:main}

To support our quantitative evaluation, we conduct additional ablation studies to analyze the effect of varying network capacity and the use of the ReLU function in our proposed functions $f_i$. More details and ablation studies are provided in the supplementary material.

\paragraph{Varying Network Capacity.}
To investigate the effect of network capacity, we vary the number of layers for the proposal network sampler of $\ourfacto$ using the Mip-NeRF 360 dataset~\cite{Barron2022MipNeRF360}. 
We report results in Tab.~\ref{tab:results_mip360_capacity_ablation} --- remarkably, lower network capacity allows for better visual quality retention for our best configuration.
As the decomposition property of NeRFs~\cite{wang2023inv} does not apply to proposal network samplers, lower network capacity forces the proposal network sampler to learn a more informative latent representation of density.
We report results for all configurations of $\ell$ and $f_i$ in the supplementary material --- the configuration $\ell=1$ and $f_2$ performs best for all tested network capacities.

\begin{table}[t]
  \footnotesize
  \setlength{\tabcolsep}{4pt}
  \centering
    \begin{tabular}{@{}crrrrrr@{}}
    \toprule
    \#Layers & \multicolumn{3}{c}{$\ourfacto$} & \multicolumn{3}{c}{Ours, $\ell = 1$, $f_2$} \\
    \cmidrule(lr){2-4}
    \cmidrule(lr){5-7}
    & PSNR $\uparrow$ & SSIM $\uparrow$ & LPIPS $\downarrow$ & PSNR $\uparrow$ & SSIM $\uparrow$ & LPIPS $\downarrow$\\
    \midrule
    2 & \ttri 24.201 & 0.726 & 0.247 & \tone 23.686 & \tone 0.704 & \tone 0.267 \\
    3 & \tone 24.272 & \ttwo 0.732 & \ttwo 0.237 & \ttwo 23.310 & \ttwo 0.700 & \ttwo 0.268 \\
    4 & 24.201 & \ttri 0.732 & \ttri 0.237 & 22.925 & 0.691 & 0.279 \\
    5 & 24.176 & 0.729 & 0.240 & \ttri 22.941 & \ttri 0.692 & \ttri 0.276 \\
    6 & \ttwo 24.247 & \tone 0.732 & \tone 0.234 & 22.891 & 0.690 & 0.278 \\
    \bottomrule
    \end{tabular}
\caption{Effect of varying network capacity for $\ourfacto$ using the Mip-NeRF 360 dataset~\cite{Barron2022MipNeRF360}: 
When we decrease capacity and use the previous best configuration of $\ell = 1$ and $f_2$, our method retains more quality and exhibits better image metrics.}
\label{tab:results_mip360_capacity_ablation}
\end{table}

\paragraph{ReLU or no ReLU.}
We also conduct an additional ablation study to test the effectiveness of applying the ReLU in Eqns.~\eqref{eq:std},~\eqref{eq:stdhalf} and~\eqref{eq:stdhalfsq}:
We remove the ReLU for all $f_i$, use the $\ourfacto$ model for the Mip-NeRF 360 dataset~\cite{Barron2022MipNeRF360} and report the results in Tab.~\ref{tab:supp:relu}.

\begin{table}[t!]
  \footnotesize
  \setlength{\tabcolsep}{4.5pt}
  \centering
    \begin{tabular}{@{}llrrrrrr@{}}
    \toprule
         && \multicolumn{3}{c}{with ReLU} & \multicolumn{3}{c}{without ReLU} \\
         \cmidrule(lr){3-5}\cmidrule(lr){6-8}
     && PSNR $\uparrow$ & SSIM $\uparrow$ & LPIPS $\downarrow$ & PSNR $\uparrow$ & SSIM $\uparrow$ & LPIPS $\downarrow$\\
    \midrule
\multirow{3}{*}{1} &$f_1$ & \ttri 22.796 & \ttri 0.687 & \ttwo 0.282 & \ttri 22.347 & \ttri 0.666 & \ttri 0.291 \\
&$f_2$ & \tone 22.925 & \tone 0.691 & \tone 0.279 & \tone 22.565 & \tone 0.671 & \tone 0.284 \\
&$f_3$ & \ttwo 22.804 & \ttwo 0.687 & \ttri 0.283 & \ttwo 22.535 & \ttwo 0.670 & \ttwo 0.285 \\\midrule
\multirow{3}{*}{2} &$f_1$ & 22.708 & 0.683 & 0.289 & 21.527 & 0.637 & 0.327 \\
&$f_2$ & 22.656 & 0.681 & 0.292 & 21.536 & 0.637 & 0.327 \\
&$f_3$ & 22.666 & 0.680 & 0.297 & 21.501 & 0.635 & 0.329 \\\midrule
\multirow{3}{*}{3} &$f_1$ & 22.443 & 0.675 & 0.305 & 21.598 & 0.635 & 0.334 \\
&$f_2$ & 22.214 & 0.669 & 0.309 & 21.618 & 0.637 & 0.332 \\
&$f_3$ & 22.390 & 0.669 & 0.316 & 21.570 & 0.636 & 0.333 \\
    \bottomrule
    \end{tabular}
  \caption{{Ablation study} on the effectiveness of using the ReLU function.
  We evaluate our method with $f_i$, either using the ReLU or not, for the Mip-NeRF 360 dataset~\cite{Barron2022MipNeRF360} and report averaged scores.
  With the ReLU, we obtain better performance for all configurations.
  }
  \label{tab:supp:relu}
\end{table}

As we can see, not applying the ReLU leads to worse outputs for every configuration we tested.
Empirically, this is caused by the more inconsistent histograms and larger standard deviation $\sigma(\fl)$.
We note that the configuration $\ell=1, f_2$ performs best for both variants of our method.

\section{Limitations}
\begin{figure}[h!]
    \centering
    \includegraphics[width=.24\linewidth]{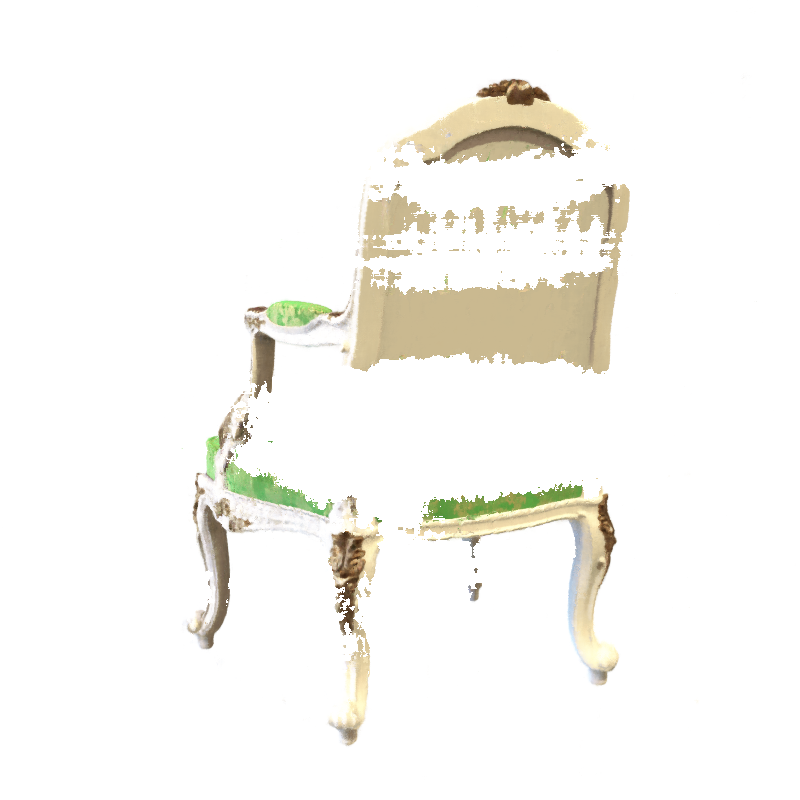}
    \includegraphics[width=.24\linewidth]{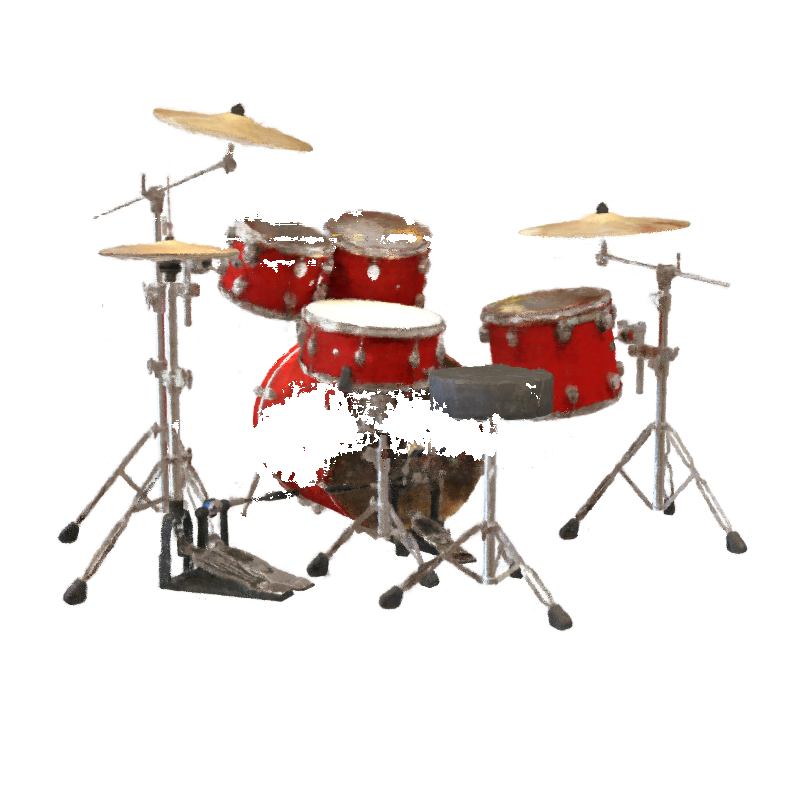}
    \includegraphics[width=.24\linewidth]{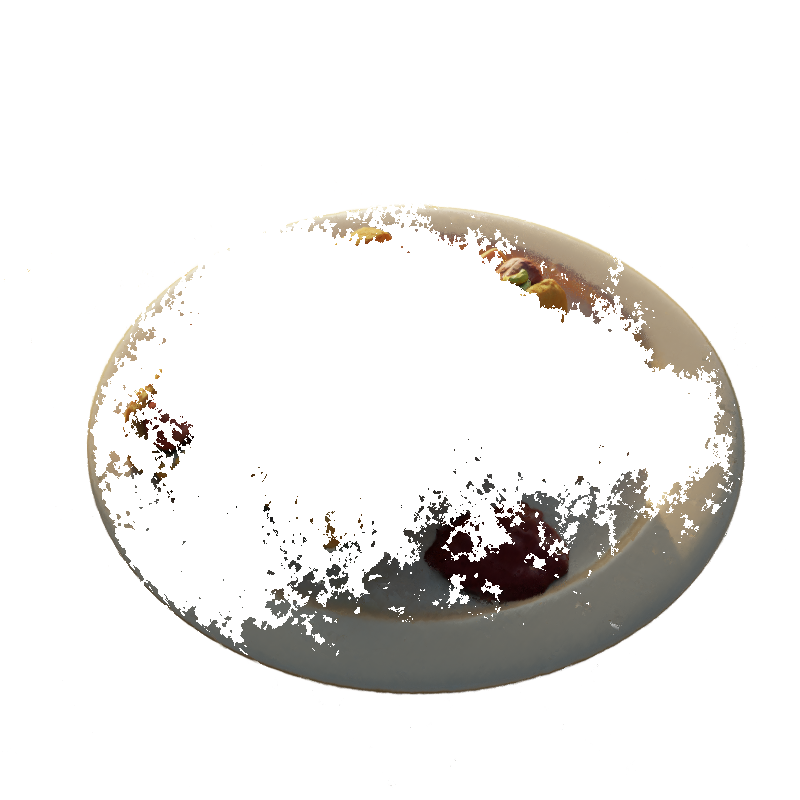}
    \includegraphics[width=.24\linewidth]{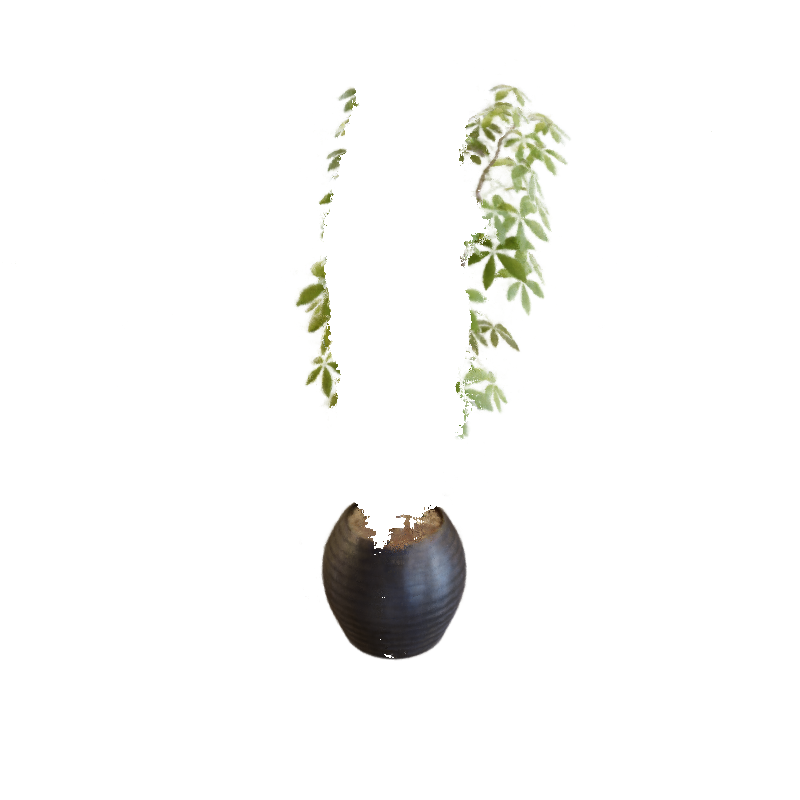}
 \begin{minipage}[t]{0.2375\linewidth}
    \centering
    \footnotesize
    \text{Chair}
\end{minipage}
 \begin{minipage}[t]{0.2375\linewidth}
    \centering
    \footnotesize
    \text{Drums}
\end{minipage}
 \begin{minipage}[t]{0.2375\linewidth}
    \centering
    \footnotesize
    \text{Hotdog}
\end{minipage}
 \begin{minipage}[t]{0.2375\linewidth}
    \centering
    \footnotesize
    \text{Ficus}
\end{minipage}
    \caption{Examples of failures of our method.
    Using Eqn.~\eqref{eq:std} may lead to rays $\vec r$ with $\hat{\vec w} = \mathbf{0}$, producing wrong samples for the subsequent fine pass.}
    \label{fig:failure}
\end{figure}

Our method is not without its limitations.
For certain test set views, using Eqn.~\eqref{eq:std} results in rays $\vec r$ with $\hat{\vec w} = \mathbf{0}$, which does not produce a good set of samples for the subsequent shading pass.
We indicate these cases in Tab.~\ref{tab:results} and show examples in Fig.~\ref{fig:failure}. 
These failures are particularly evident for Mip-NeRF, as the fine pass does not include the uniformly distributed samples.
Eqns.~\eqref{eq:stdhalf}~and~\eqref{eq:stdhalfsq} mitigate this issue effectively, but might lead to outputs with less detail due to their more conservative formulation.

Although our approach performs well for a variety of architectures and datasets, extending it to hybrid or explicit representations such as~\cite{YuPlenoxels, MuellerNGP, dvgo, pointnerf} is not trivial, as these approaches replace the MLP with more efficient alternatives.
Further, our handcrafted functions for density estimation were created intuitively and are biased towards uniform or piecewise-linear sample placement.

\section{Conclusion and Outlook}
We have presented a generally applicable framework to analyze and visualize the intermediate activations of coordinate-based ReLU-MLPs. 
Building on our findings, we proposed a novel method to obtain density estimates using our derived activation features.
We demonstrated that our approach produces faithful renderings with improved performance and provided valuable insight into the internals of neural fields.
We believe our concepts could be applied to other tasks involving coordinate-based MLPs.
{
    \small
    \bibliographystyle{ieeenat_fullname}
    \bibliography{sample}

\begin{thebibliography}{48}
\providecommand{\natexlab}[1]{#1}
\providecommand{\url}[1]{\texttt{#1}}
\expandafter\ifx\csname urlstyle\endcsname\relax
  \providecommand{\doi}[1]{doi: #1}\else
  \providecommand{\doi}{doi: \begingroup \urlstyle{rm}\Url}\fi

\bibitem[Arandjelovi{\'c} and Zisserman(2021)]{arandjelovic2021nerfindetail}
Relja Arandjelovi{\'c} and Andrew Zisserman.
\newblock {NeRF in detail: Learning to sample for view synthesis}.
\newblock \emph{arXiv CoRR}, abs/2106.05264, 2021.

\bibitem[Barron et~al.(2021)Barron, Mildenhall, Tancik, Hedman, Martin-Brualla, and Srinivasan]{barron2021mipnerf}
Jonathan~T. Barron, Ben Mildenhall, Matthew Tancik, Peter Hedman, Ricardo Martin-Brualla, and Pratul~P. Srinivasan.
\newblock {Mip-NeRF: A Multiscale Representation for Anti-Aliasing Neural Radiance Fields}.
\newblock In \emph{ICCV}, 2021.

\bibitem[Barron et~al.(2022)Barron, Mildenhall, Verbin, Srinivasan, and Hedman]{Barron2022MipNeRF360}
Jonathan~T. Barron, Ben Mildenhall, Dor Verbin, Pratul~P. Srinivasan, and Peter Hedman.
\newblock {Mip-NeRF 360: Unbounded Anti-Aliased Neural Radiance Fields}.
\newblock In \emph{CVPR}, 2022.

\bibitem[Chen et~al.(2022)Chen, Xu, Geiger, Yu, and Su]{TensorRFECCV}
Anpei Chen, Zexiang Xu, Andreas Geiger, Jingyi Yu, and Hao Su.
\newblock {TensoRF: Tensorial Radiance Fields}.
\newblock In \emph{ECCV}, 2022.

\bibitem[Chen et~al.(2023)Chen, Funkhouser, Hedman, and Tagliasacchi]{chen2022mobilenerf}
Zhiqin Chen, Thomas Funkhouser, Peter Hedman, and Andrea Tagliasacchi.
\newblock {MobileNeRF: Exploiting the Polygon Rasterization Pipeline for Efficient Neural Field Rendering on Mobile Architectures}.
\newblock In \emph{CVPR}, 2023.

\bibitem[Chng et~al.(2022)Chng, Ramasinghe, Sherrah, and Lucey]{chng2022gaussian}
Shin-Fang Chng, Sameera Ramasinghe, Jamie Sherrah, and Simon Lucey.
\newblock {Gaussian Activated Neural Radiance Fields for High Fidelity Reconstruction \& Pose Estimation}.
\newblock In \emph{ECCV}, 2022.

\bibitem[Fathony et~al.(2021)Fathony, Sahu, Willmott, and Kolter]{fathony2020multiplicative}
Rizal Fathony, Anit~Kumar Sahu, Devin Willmott, and J~Zico Kolter.
\newblock {Multiplicative Filter Networks}.
\newblock In \emph{ICLR}, 2021.

\bibitem[Fridovich-Keil et~al.(2022)Fridovich-Keil, Yu, Tancik, Chen, Recht, and Kanazawa]{YuPlenoxels}
Sara Fridovich-Keil, Alex Yu, Matthew Tancik, Qinhong Chen, Benjamin Recht, and Angjoo Kanazawa.
\newblock {Plenoxels: Radiance Fields without Neural Networks}.
\newblock In \emph{CVPR}, 2022.

\bibitem[Garbin et~al.(2021)Garbin, Kowalski, Johnson, Shotton, and Valentin]{GarbinFastNeRF}
Stephan~J. Garbin, Marek Kowalski, Matthew Johnson, Jamie Shotton, and Julien Valentin.
\newblock {FastNeRF: High-Fidelity Neural Rendering at 200FPS}.
\newblock In \emph{ICCV}, 2021.

\bibitem[Hedman et~al.(2018)Hedman, Philip, Price, Frahm, Drettakis, and Brostow]{DeepBlending2018}
Peter Hedman, Julien Philip, True Price, Jan-Michael Frahm, George Drettakis, and Gabriel Brostow.
\newblock {Deep Blending for Free-viewpoint Image-based Rendering}.
\newblock \emph{ACM TOG}, 37\penalty0 (6), 2018.

\bibitem[Hedman et~al.(2021)Hedman, Srinivasan, Mildenhall, Barron, and Debevec]{hedmanBaking}
Peter Hedman, Pratul~P. Srinivasan, Ben Mildenhall, Jonathan~T. Barron, and Paul~E. Debevec.
\newblock {Baking Neural Radiance Fields for Real-Time View Synthesis}.
\newblock In \emph{ICCV}, 2021.

\bibitem[Hu et~al.(2022)Hu, Liu, Chen, Shen, and Jia]{HuEfficient}
Tao Hu, Shu Liu, Yilun Chen, Tiancheng Shen, and Jiaya Jia.
\newblock {EfficientNeRF: Efficient Neural Radiance Fields}.
\newblock In \emph{CVPR}, 2022.

\bibitem[Karnewar et~al.(2022)Karnewar, Ritschel, Wang, and Mitra]{Karnewar2022ReLUFields}
Animesh Karnewar, Tobias Ritschel, Oliver Wang, and Niloy~J. Mitra.
\newblock {ReLU Fields: The Little Non-linearity That Could}.
\newblock In \emph{SIGGRAPH}, 2022.

\bibitem[Kerbl et~al.(2023)Kerbl, Kopanas, Leimk{\"u}hler, and Drettakis]{kerbl3Dgaussians}
Bernhard Kerbl, Georgios Kopanas, Thomas Leimk{\"u}hler, and George Drettakis.
\newblock {3D Gaussian Splatting for Real-Time Radiance Field Rendering}.
\newblock \emph{ACM TOG}, 42\penalty0 (4), 2023.

\bibitem[Kulhanek and Sattler(2023)]{TetraNeRF}
Jonas Kulhanek and Torsten Sattler.
\newblock {Tetra-NeRF: Representing Neural Radiance Fields Using Tetrahedra}.
\newblock In \emph{ICCV}, 2023.

\bibitem[Kurz et~al.(2022)Kurz, Neff, Lv, Zollh\"{o}fer, and Steinberger]{KurzAdaNeRF}
Andreas Kurz, Thomas Neff, Zhaoyang Lv, Michael Zollh\"{o}fer, and Markus Steinberger.
\newblock {AdaNeRF: Adaptive Sampling for Real-time Rendering of Neural Radiance Fields}.
\newblock In \emph{ECCV}, 2022.

\bibitem[Lin et~al.(2022)Lin, Peng, Xu, Yan, Shuai, Bao, and Zhou]{lin2022efficient}
Haotong Lin, Sida Peng, Zhen Xu, Yunzhi Yan, Qing Shuai, Hujun Bao, and Xiaowei Zhou.
\newblock {Efficient Neural Radiance Fields with Learned Depth-Guided Sampling}.
\newblock In \emph{SIGGRAPH Asia}, 2022.

\bibitem[Martin-Brualla et~al.(2021)Martin-Brualla, Radwan, Sajjadi, Barron, Dosovitskiy, and Duckworth]{martinbrualla2020nerfw}
Ricardo Martin-Brualla, Noha Radwan, Mehdi S.~M. Sajjadi, Jonathan~T. Barron, Alexey Dosovitskiy, and Daniel Duckworth.
\newblock {NeRF in the Wild: Neural Radiance Fields for Unconstrained Photo Collections}.
\newblock In \emph{CVPR}, 2021.

\bibitem[Mildenhall et~al.(2019)Mildenhall, Srinivasan, Ortiz-Cayon, Kalantari, Ramamoorthi, Ng, and Kar]{mildenhall2019llff}
Ben Mildenhall, Pratul~P. Srinivasan, Rodrigo Ortiz-Cayon, Nima~Khademi Kalantari, Ravi Ramamoorthi, Ren Ng, and Abhishek Kar.
\newblock {Local Light Field Fusion: Practical View Synthesis with Prescriptive Sampling Guidelines}.
\newblock \emph{ACM TOG}, 38\penalty0 (4), 2019.

\bibitem[Mildenhall et~al.(2020)Mildenhall, Srinivasan, Tancik, Barron, Ramamoorthi, and Ng]{MildenhallNeRF}
Ben Mildenhall, Pratul~P. Srinivasan, Matthew Tancik, Jonathan~T. Barron, Ravi Ramamoorthi, and Ren Ng.
\newblock {NeRF: Representing Scenes as Neural Radiance Fields for View Synthesis}.
\newblock In \emph{ECCV}, 2020.

\bibitem[M\"uller et~al.(2022)M\"uller, Evans, Schied, and Keller]{MuellerNGP}
Thomas M\"uller, Alex Evans, Christoph Schied, and Alexander Keller.
\newblock {Instant Neural Graphics Primitives with a Multiresolution Hash Encoding}.
\newblock \emph{ACM TOG}, 41\penalty0 (4), 2022.

\bibitem[Neff et~al.(2021)Neff, Stadlbauer, Parger, Kurz, Mueller, Chaitanya, Kaplanyan, and Steinberger]{NeffDONeRF}
Thomas Neff, Pascal Stadlbauer, Mathias Parger, Andreas Kurz, Joerg~H. Mueller, Chakravarty R.~Alla Chaitanya, Anton~S. Kaplanyan, and Markus Steinberger.
\newblock {DONeRF: Towards Real-Time Rendering of Compact Neural Radiance Fields using Depth Oracle Networks}.
\newblock \emph{Comput. Graph. Forum}, 40\penalty0 (4):\penalty0 45--59, 2021.

\bibitem[Oechsle et~al.(2021)Oechsle, Peng, and Geiger]{OechsleUnisurf}
Michael Oechsle, Songyou Peng, and Andreas Geiger.
\newblock {UNISURF: Unifying Neural Implicit Surfaces and Radiance Fields for Multi-View Reconstruction}.
\newblock In \emph{ICCV}, 2021.

\bibitem[Piala and Clark({2021})]{PialaTerminerf}
Martin Piala and Ronald Clark.
\newblock {TermiNeRF: Ray Termination Prediction for Efficient Neural Rendering}.
\newblock In \emph{3DV}, {2021}.

\bibitem[Rahaman et~al.(2019)Rahaman, Baratin, Arpit, Draxler, Lin, Hamprecht, Bengio, and Courville]{rahamanSpectral}
Nasim Rahaman, Aristide Baratin, Devansh Arpit, Felix Draxler, Min Lin, Fred Hamprecht, Yoshua Bengio, and Aaron Courville.
\newblock {On the Spectral Bias of Neural Networks}.
\newblock In \emph{ICLR}, 2019.

\bibitem[Ramasinghe and Lucey(2022)]{ramasinghe2022beyond}
Sameera Ramasinghe and Simon Lucey.
\newblock {Beyond Periodicity: Towards a Unifying Framework for Activations in Coordinate-MLPs}.
\newblock In \emph{ECCV}, 2022.

\bibitem[Rebain et~al.(2021)Rebain, Jiang, Yazdani, Li, Yi, and Tagliasacchi]{rebainDerf}
Daniel Rebain, Wei Jiang, Soroosh Yazdani, Ke Li, Kwang~M. Yi, and Andrea Tagliasacchi.
\newblock {DeRF: Decomposed Radiance Fields}.
\newblock In \emph{CVPR}, 2021.

\bibitem[Reiser et~al.(2021)Reiser, Peng, Liao, and Geiger]{ReiserKiloNeRF}
Christian Reiser, Songyou Peng, Yiyi Liao, and Andreas Geiger.
\newblock {KiloNeRF: Speeding up Neural Radiance Fields with Thousands of Tiny MLPs}.
\newblock In \emph{ICCV}, 2021.

\bibitem[Reiser et~al.(2024)Reiser, Garbin, Srinivasan, Verbin, Szeliski, Mildenhall, Barron, Hedman, and Geiger]{reiser2024binary}
Christian Reiser, Stephan Garbin, Pratul~P Srinivasan, Dor Verbin, Richard Szeliski, Ben Mildenhall, Jonathan~T Barron, Peter Hedman, and Andreas Geiger.
\newblock {Binary Opacity Grids: Capturing Fine Geometric Detail for Mesh-Based View Synthesis}.
\newblock \emph{arXiv CoRR}, abs/2402.12377, 2024.

\bibitem[Saragadam et~al.(2023)Saragadam, LeJeune, Tan, Balakrishnan, Veeraraghavan, and Baraniuk]{saragadam2023wire}
Vishwanath Saragadam, Daniel LeJeune, Jasper Tan, Guha Balakrishnan, Ashok Veeraraghavan, and Richard~G Baraniuk.
\newblock {WIRE: Wavelet Implicit Neural Representations}.
\newblock In \emph{CVPR}, 2023.

\bibitem[Sitzmann et~al.(2020)Sitzmann, Martel, Bergman, Lindell, and Wetzstein]{sitzmannSiren}
Vincent Sitzmann, Julien~N.P. Martel, Alexander~W. Bergman, David~B. Lindell, and Gordon Wetzstein.
\newblock {Implicit Neural Representations with Periodic Activation Functions}.
\newblock In \emph{NeurIPS}, 2020.

\bibitem[Springenberg et~al.(2015)Springenberg, Dosovitskiy, Brox, and Riedmiller]{SpringenbergAllConv}
Jost~T. Springenberg, Alexey Dosovitskiy, Thomas Brox, and Martin Riedmiller.
\newblock {Striving for Simplicity: The All Convolutional Net}.
\newblock In \emph{ICLR}, 2015.

\bibitem[Sun et~al.(2022)Sun, Sun, and Chen]{dvgo}
Cheng Sun, Min Sun, and Hwann{-}Tzong Chen.
\newblock {Direct Voxel Grid Optimization: Super-fast Convergence for Radiance Fields Reconstruction}.
\newblock In \emph{CVPR}, 2022.

\bibitem[Tancik et~al.(2020)Tancik, Srinivasan, Mildenhall, Fridovich-Keil, Raghavan, Singhal, Ramamoorthi, Barron, and Ng]{TancikFourier}
Matthew Tancik, Pratul~P. Srinivasan, Ben Mildenhall, Sara Fridovich-Keil, Nithin Raghavan, Utkarsh Singhal, Ravi Ramamoorthi, Jonathan~T. Barron, and Ren Ng.
\newblock {Fourier Features Let Networks Learn High Frequency Functions in Low Dimensional Domains}.
\newblock In \emph{NeurIPS}, 2020.

\bibitem[Tancik et~al.(2023)Tancik, Weber, Ng, Li, Yi, Kerr, Wang, Kristoffersen, Austin, Salahi, Ahuja, McAllister, and Kanazawa]{tancik2023nerfstudio}
Matthew Tancik, Ethan Weber, Evonne Ng, Ruilong Li, Brent Yi, Justin Kerr, Terrance Wang, Alexander Kristoffersen, Jake Austin, Kamyar Salahi, Abhik Ahuja, David McAllister, and Angjoo Kanazawa.
\newblock {Nerfstudio: A Modular Framework for Neural Radiance Field Development}.
\newblock In \emph{SIGGRAPH}, 2023.

\bibitem[Verbin et~al.(2022)Verbin, Hedman, Mildenhall, Zickler, Barron, and Srinivasan]{verbin2022refnerf}
Dor Verbin, Peter Hedman, Ben Mildenhall, Todd Zickler, Jonathan~T. Barron, and Pratul~P. Srinivasan.
\newblock {Ref-NeRF: Structured View-Dependent Appearance for Neural Radiance Fields}.
\newblock In \emph{CVPR}, 2022.

\bibitem[Wan et~al.(2023)Wan, Richardt, Bo\v{z}i\v{c}, Li, Rengarajan, Nam, Xiang, Li, Zhu, Ranjan, and Liao]{DuplexNeRF}
Ziyu Wan, Christian Richardt, Alja\v{z} Bo\v{z}i\v{c}, Chao Li, Vijay Rengarajan, Seonghyeon Nam, Xiaoyu Xiang, Tuotuo Li, Bo Zhu, Rakesh Ranjan, and Jing Liao.
\newblock {Learning Neural Duplex Radiance Fields for Real-Time View Synthesis}.
\newblock In \emph{CVPR}, 2023.

\bibitem[Wang et~al.(2023{\natexlab{a}})Wang, Supikov, Ratcliff, Fuchs, and Azuma]{wang2023inv}
Shengze Wang, Alexey Supikov, Joshua Ratcliff, Henry Fuchs, and Ronald Azuma.
\newblock {INV: Towards Streaming Incremental Neural Videos}.
\newblock \emph{arXiv CoRR}, abs/2302.01532, 2023{\natexlab{a}}.

\bibitem[Wang et~al.(2004)Wang, Bovik, Sheikh, and Simoncelli]{WangSSIM}
Zhou Wang, Alam~C. Bovik, Hamid~R. Sheikh, and Eero~P. Simoncelli.
\newblock {Image Quality Assessment: From Error Visibility to Structural Similarity}.
\newblock \emph{IEEE TIP}, 13\penalty0 (4):\penalty0 600--612, 2004.

\bibitem[Wang et~al.(2021)Wang, Wu, Xie, Chen, and Prisacariu]{wang2021nerfmm}
Zirui Wang, Shangzhe Wu, Weidi Xie, Min Chen, and Victor~Adrian Prisacariu.
\newblock {NeRF$--$: Neural Radiance Fields Without Known Camera Parameters}.
\newblock \emph{arXiv CoRR}, abs/2102.07064, 2021.

\bibitem[Wang et~al.(2023{\natexlab{b}})Wang, Shen, Nimier-David, Sharp, Gao, Keller, Fidler, M\"uller, and Gojcic]{adaptiveshells2023}
Zian Wang, Tianchang Shen, Merlin Nimier-David, Nicholas Sharp, Jun Gao, Alexander Keller, Sanja Fidler, Thomas M\"uller, and Zan Gojcic.
\newblock {Adaptive Shells for Efficient Neural Radiance Field Rendering}.
\newblock \emph{ACM TOG}, 42\penalty0 (6), 2023{\natexlab{b}}.

\bibitem[Wu et~al.(2022)Wu, Lee, Bhattad, Wang, and Forsyth]{diver}
Liwen Wu, Jae~Yong Lee, Anand Bhattad, Yuxiong Wang, and David Forsyth.
\newblock {DIVeR: Real-time and Accurate Neural Radiance Fields with Deterministic Integration for Volume Rendering}.
\newblock In \emph{CVPR}, 2022.

\bibitem[Xie et~al.(2022)Xie, Takikawa, Saito, Litany, Yan, Khan, Tombari, Tompkin, Sitzmann, and Sridhar]{XieNeuralFields}
Yiheng Xie, Towaki Takikawa, Shunsuke Saito, Or Litany, Shiqin Yan, Numair Khan, Federico Tombari, James Tompkin, Vincent Sitzmann, and Srinath Sridhar.
\newblock {Neural Fields in Visual Computing and Beyond}.
\newblock \emph{Comput. Graph. Forum}, 41\penalty0 (2):\penalty0 641--676, 2022.

\bibitem[Xu et~al.(2022)Xu, Xu, Philip, Bi, Shu, Sunkavalli, and Neumann]{pointnerf}
Qiangeng Xu, Zexiang Xu, Julien Philip, Sai Bi, Zhixin Shu, Kalyan Sunkavalli, and Ulrich Neumann.
\newblock {Point-NeRF: Point-based Neural Radiance Fields}.
\newblock In \emph{CVPR}, 2022.

\bibitem[Yariv et~al.(2023)Yariv, Hedman, Reiser, Verbin, Srinivasan, Szeliski, Barron, and Mildenhall]{yariv2023bakedsdf}
Lior Yariv, Peter Hedman, Christian Reiser, Dor Verbin, Pratul~P Srinivasan, Richard Szeliski, Jonathan~T Barron, and Ben Mildenhall.
\newblock {BakedSDF: Meshing Neural SDFs for Real-Time View Synthesis}.
\newblock In \emph{SIGGRAPH}, 2023.

\bibitem[Yen-Chen(2020)]{lin2020nerfpytorch}
Lin Yen-Chen.
\newblock {NeRF-pytorch}.
\newblock \url{https://github.com/yenchenlin/nerf-pytorch/}, 2020.

\bibitem[Zeiler and Fergus(2014)]{ZeilerVisualizingCNN}
Matthew~D. Zeiler and Rob Fergus.
\newblock {Visualizing and Understanding Convolutional Networks}.
\newblock In \emph{ECCV}, 2014.

\bibitem[Zhang et~al.(2018)Zhang, Isola, Efros, Shechtman, and Wang]{ZhangLPIPS}
Richard Zhang, Phillip Isola, Alexei~A. Efros, Eli Shechtman, and Oliver Wang.
\newblock {The Unreasonable Effectiveness of Deep Features as a Perceptual Metric}.
\newblock In \emph{CVPR}, 2018.

\end{thebibliography}
}

\clearpage
\appendix
\maketitlesupplementary

\section{Additional Implementation Details for nerfacto$^\dagger$}
\paragraph{Blender.}
For the Blender dataset~\cite{MildenhallNeRF}, we set $t_n = 2.0,\ t_f = 6.0$ for all scenes.
We disable the camera optimizer, the appearance embedding, the distortion loss and the scene contraction.
We manually set the background color to white.
For all scenes, we use a batch size of $2048$ rays during training and use a uniform initial proposal network sampler.
We train $\text{nerfacto}^\dagger$ for $16.5 \times 10^{3}$ epochs, as in~\cite{tancik2023nerfstudio}.

\paragraph{LLFF.}
For the LLFF dataset~\cite{mildenhall2019llff}, we disable the camera optimizer, the appearance embedding and the distortion loss. 
We use a $\ell_\infty$ scene contraction as default for nerfacto~\cite{tancik2023nerfstudio}, contrary to the use of NDC coordinates in NeRF and Mip-NeRF.
We do not use our custom LLFF dataparser, we instead opt for the nerfstudio dataparser, which requires pre-processing the data.
For all scenes, we use a batch size of $1024$ rays during training and a batch size of $2048$ during evaluation.
We train $\text{nerfacto}^\dagger$ for $30 \times 10^{3}$ epochs, as in~\cite{tancik2023nerfstudio}.
For the \emph{leaves} scenes, we use a uniform initial proposal network sampler, as the piecewise-uniform sampler did not result in stable optimization.

\paragraph{Mip-NeRF 360 and Deep Blending.}
For Mip-NeRF 360~\cite{Barron2022MipNeRF360} and Deep Blending~\cite{DeepBlending2018}, we use the Nerfstudio data processing pipeline~\cite{tancik2023nerfstudio} to extract the dataset using the given SfM output.
As we use an MLP with NeRFs positional encoding~\cite{MildenhallNeRF}, the hyperparameters reported in~\cite{tancik2023nerfstudio} do not necessarily provide the best results.
We experimented with different hyperparameters and ultimately used the configuration reported in the main paper, as it resulted in the best performance for $\ourfacto$.
Also, note that the results for Mip-NeRF 360 reported in~\cite{tancik2023nerfstudio} do not include the scenes \emph{flowers} and \emph{treehill}, for which $\ourfacto$ exhibits the worst image metrics (c.f. Sec.~\ref{sec:supp:per-scene}).

\paragraph{Comparison with nerfacto.}
To investigate the effect of the MLP introduced for $\ourfacto$, we compare the results with nerfacto (with \mbox{\textit{HashMLPDensityField}}) in Tab.~\ref{tab:supp:nerfacto}.
We use the Mip-NeRF 360 dataset~\cite{Barron2022MipNeRF360} and use the same hyperparameter settings as reported in the main paper.
Both variants achieve comparable results, although nerfacto is significantly faster.

\begin{table}[h!]
  \footnotesize
  \centering
    \begin{tabular}{lcccc}
    \toprule
    Method & \# Layers & PSNR $\uparrow$ & SSIM $\uparrow$ & LPIPS $\downarrow$\\
    \midrule
    &2 & 24.201 & 0.726 & 0.247 \\
    &3 & \ttwo 24.272 & \ttri 0.732 & \ttri 0.237 \\
    $\ourfacto$&4 & 24.201 & 0.732 & 0.237 \\
    &5 & 24.176 & 0.729 & 0.240 \\
    &6 & \ttri 24.247 & \ttwo 0.732 & \ttwo 0.234 \\\midrule
    \multicolumn{2}{c}{nerfacto} & \tone 24.682 & \tone 0.735 & \tone 0.229 \\
    \bottomrule
    \end{tabular}
\caption{Comparison of $\ourfacto$ (with MLP) and nerfacto (with \mbox{\textit{HashMLPDensityField}}) for the Mip-NeRF 360 dataset~\cite{Barron2022MipNeRF360}.}
\label{tab:supp:nerfacto}
\end{table}

\section{Ablation Studies}
First, we present an extensions to our method, which reduces processing time even further.
In addition, we also report more detailed results for the ablation studies in Sec.~\ref{sec:ablation:main}.

\begin{figure}[ht!]
\centering
\rotatebox{90}{
 \begin{minipage}[c]{0.1\linewidth}
    \centering
    \footnotesize
    \text{with ReLU}
\end{minipage}}
    \includegraphics[width=.45\linewidth]{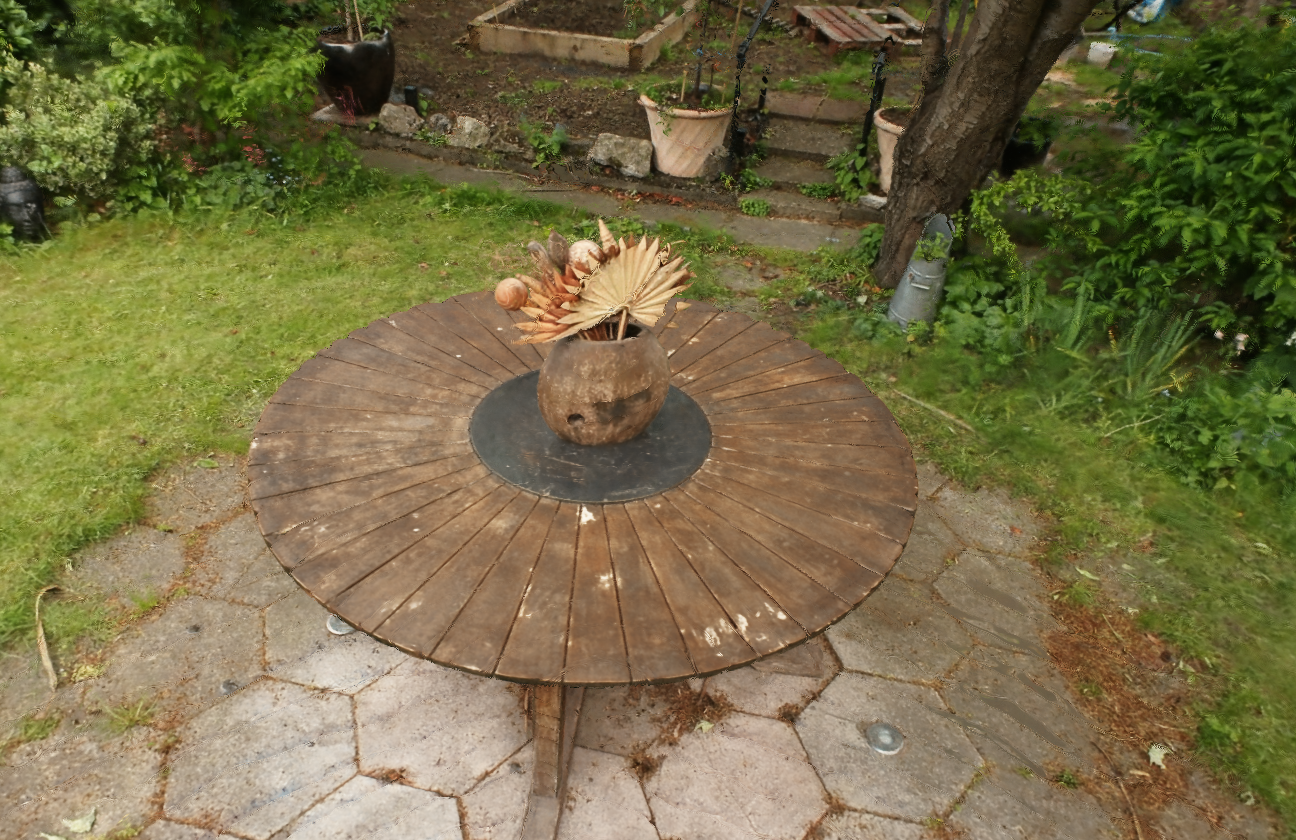}
    \includegraphics[width=.45\linewidth]{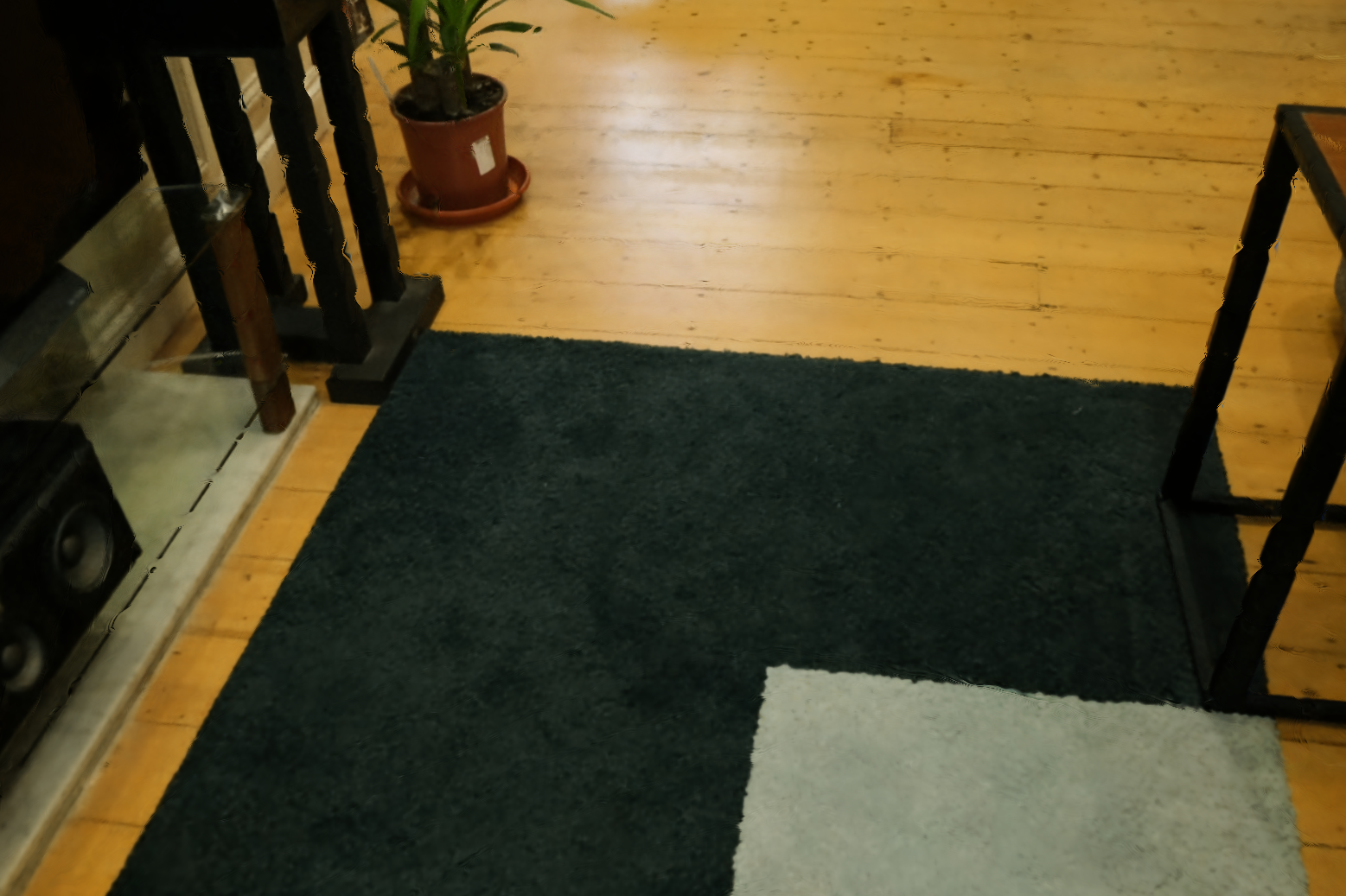}\\
\rotatebox{90}{
 \begin{minipage}[c]{0.1\linewidth}
    \centering
    \footnotesize
    \text{without ReLU}
\end{minipage}}
    \includegraphics[width=.45\linewidth]{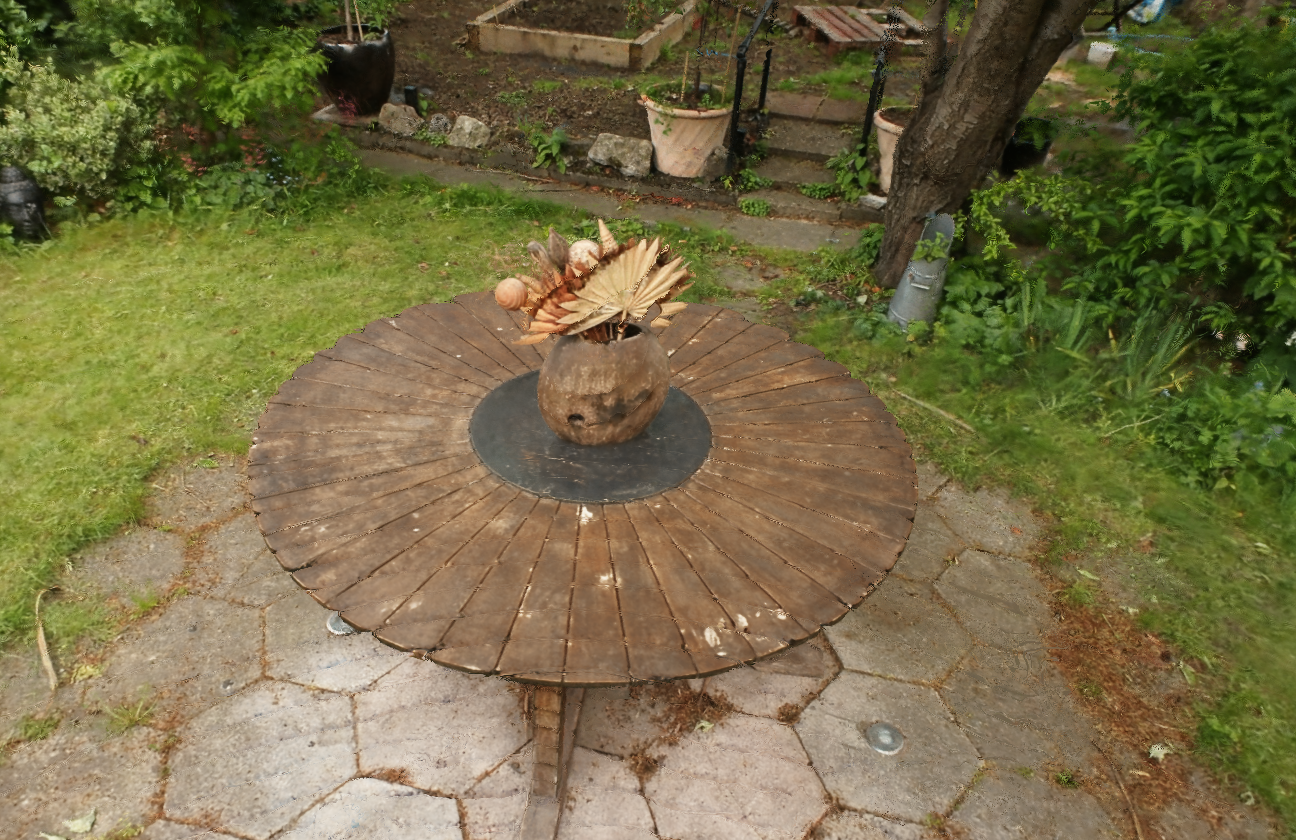}
    \includegraphics[width=.45\linewidth]{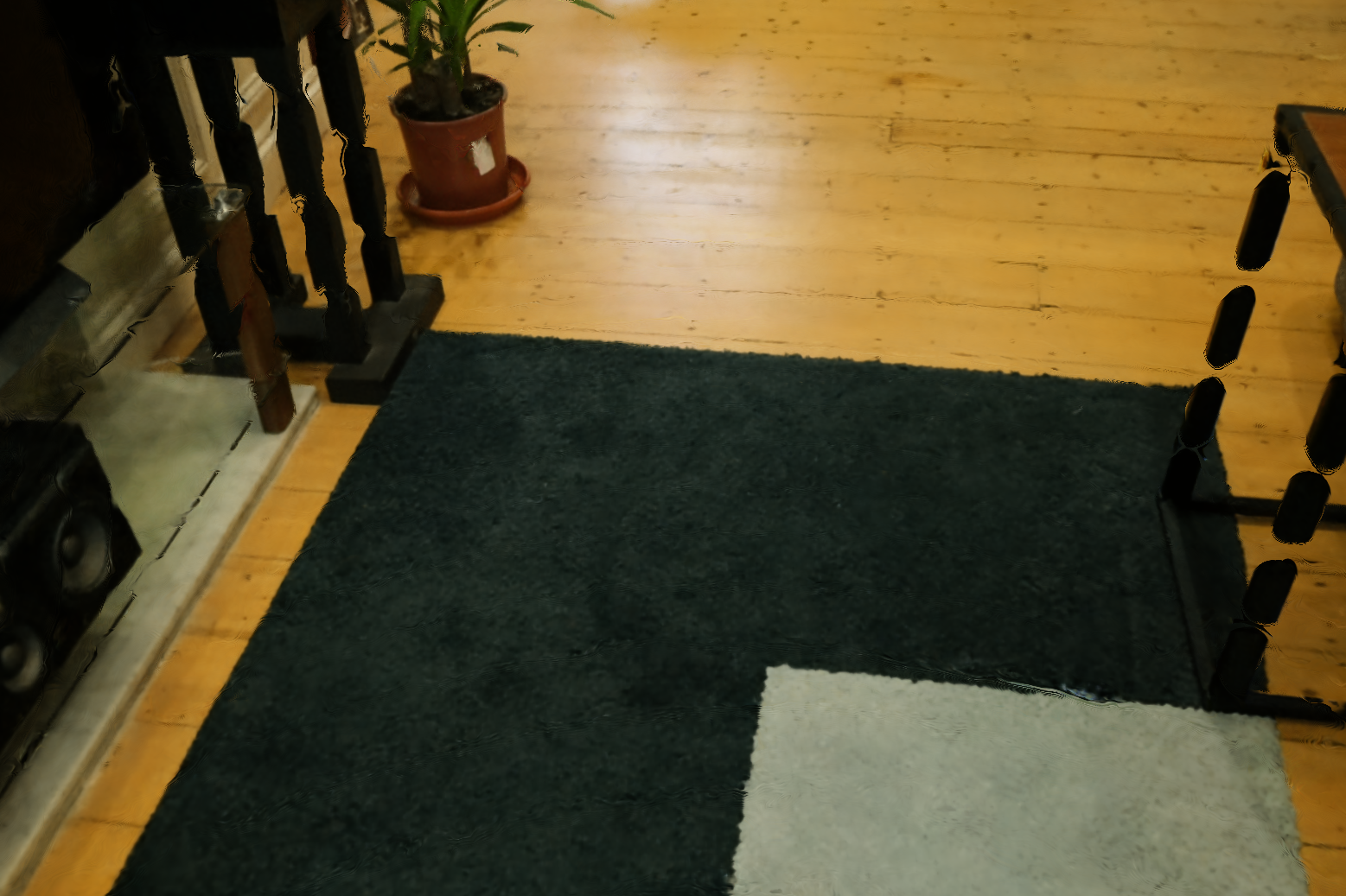}
    \caption{\textbf{Example renderings} from our our method (with ReLU) compared to without ReLU for the the Mip-NeRF 360 dataset~\cite{Barron2022MipNeRF360}.
    Not using the ReLU results in artefacts for various renderings due to not being able to capture fine geometry.}
    \label{fig:relunorelu}
\end{figure}

\begin{table}[ht!]
  \footnotesize
  \setlength{\tabcolsep}{5pt}
  \centering
    \begin{tabular}{@{}lcccccc@{}}\toprule
    \multicolumn{7}{c}{\textbf{NeRF}} \\\midrule
     \multirow{3}{*}{Method}  & \multicolumn{3}{c}{{Blender}, $800 \times 800$} & \multicolumn{3}{c}{{LLFF}, $1008 \times 756$}\\
    \cmidrule(lr){2-4}\cmidrule(lr){5-7}
     & PSNR & SSIM & LPIPS & PSNR & SSIM & LPIPS \\\midrule
     Base & 29.05 & 0.93 & 0.07 
                & 25.12 & 0.74 & 0.24\\ 
     Upsampling & 28.90 & 0.93 & 0.08
                & 24.60 & 0.73 & 0.26\\\midrule
    NeRF        &  29.87 &  0.94 &  0.06 
                                    &  25.89 &  0.78 &  0.18  \\
    \midrule
    \multicolumn{7}{c}{{$\textbf{Mip-NeRF}$}} \\\midrule
   Base & 29.04 & 0.93 & 0.07 & 25.44 & 0.75 & 0.22  \\
   Upsampling & 28.94 & 0.93 & 0.08 & 24.90& 0.74 & 0.25\\\midrule
    Mip-NeRF   &  29.69 & 0.94 & 0.06
                                    &  25.99 &  0.78 &  0.18  \\
    \bottomrule
    \end{tabular}
      \caption{\textbf{Ablation study}: We can further improve the performance of our implementation by obtaining the activations at a lower resolution.
    We use the Blender~\cite{MildenhallNeRF} and LLFF~\cite{mildenhall2019llff} datasets and use $\ell = 2$ and $f_2$, the configuration which performed best in our experiments for these datasets.}
  \label{tab:supp:ablations}
\end{table}

\subsection{Upsampling Activations}
As our approach builds on the intermediate activations for a proposal of density along a ray, a natural extension to our method is to obtain the activations $\vec A^{(\ell)}$ at a lower resolution.
To this end, we perform an ablation study, where we record $\vec A^{(\ell)}$ at a resolution of $\frac{w}{2} \times \frac{h}{2} \times \frac{N_c}{2} \times N_h$, then perform upsampling to obtain a full-resolution activation $\vec A^{(\ell)}$.
As our interpolation method, we choose nearest neighbor upsampling due to its negligible cost.
Similarly to this approach, DONeRF~\cite{NeffDONeRF} filters along the depth axis and across neighboring rays to obtain a smoothed depth target for their oracle network.
For the Blender dataset~\cite{MildenhallNeRF} and the LLFF dataset~\cite{mildenhall2019llff}, we use our method with $\ell =1$ and $f_2$ and report results in Tab.~\ref{tab:supp:ablations}.
We perform this experiments for NeRF~\cite{MildenhallNeRF} and Mip-NeRF~\cite{barron2021mipnerf}: 
As we can see, this modifications impacts visual quality only slightly.


\begin{table}[ht!]
  \scriptsize
  \setlength{\tabcolsep}{2.7pt}
  \centering
    \begin{tabular}{@{}llrrrrrrrr@{}}
    \toprule
         && \multicolumn{4}{c}{Reduced Capacity: 2 Layers} & \multicolumn{4}{c}{Reduced Capacity: 3 Layers} \\
         \cmidrule(lr){3-6}\cmidrule(lr){7-10}
     && PSNR & SSIM & LPIPS & Speedup & PSNR & SSIM & LPIPS & Speedup\\
    \midrule
\multirow{3}{*}{1} &$f_1$ & \ttri 23.66 & \ttri 0.703 & \ttri 0.267 & 
        & 23.20 & 0.696 & \ttri 0.272 &\\
&$f_2$ & \ttwo 23.69 & \ttwo 0.704 & \ttwo 0.267 & -
        & \ttwo 23.31 & \ttwo 0.700 & \ttwo 0.268 & 44\%  \\
& $f_3$ & 23.66 & 0.703 & 0.269 &  
        & \ttri 23.20 & \ttri 0.697 & 0.272 &   \\
    \midrule
\multirow{3}{*}{2} &$f_1$ & &  &  & 
    & 22.63 & 0.677 & 0.299 &\\
&$f_2$ &  &  &  & 
    & 22.56 & 0.676 & 0.299 & -\\
&$f_3$ &  &  &  & 
    & 22.71 & 0.676 & 0.306 &\\\midrule
    \multicolumn{2}{c}{$\text{nerfacto}^\dagger$} & \tone 24.20 & \tone 0.726 & \tone 0.247 & -  & \tone 24.27 & \tone 0.732 & \tone 0.237 & -\\
    \bottomrule
    \\[-.2ex]
    \\[-.2ex]
    \toprule
         && \multicolumn{4}{c}{Increased Capacity: 5 Layers} & \multicolumn{4}{c}{Increased Capacity: 6 Layers} \\
         \cmidrule(lr){3-6}\cmidrule(lr){7-10}
     && PSNR & SSIM & LPIPS & Speedup & PSNR & SSIM & LPIPS & Speedup\\
    \midrule
    \multirow{3}{*}{1} &$f_1$ & 22.83 & \ttri 0.689 & \ttri 0.279 &
& 22.73 & \ttri 0.686 & \ttri 0.282 & \\
    &$f_2$ & \ttwo 22.94 & \ttwo 0.692 & \ttwo 0.276 & 101\% 
& \ttwo 22.89 & \ttwo 0.690 & \ttwo 0.278 & 120\%\\
    & $f_3$ & 22.83 & 0.688 & 0.280 &  
& \ttri 22.75 & 0.686 & 0.283 &\\
    \midrule
    \multirow{3}{*}{2} &$f_1$ & 22.82 & 0.688 & 0.280 & 
& 22.67 & 0.683 & 0.287 &\\
    &$f_2$ & \ttri 22.84 & 0.688 & 0.281 & 50\% 
& 22.69 & 0.683 & 0.289 & 69\%\\
     &$f_3$ & 22.78 & 0.686 & 0.284 & 
& 22.62 & 0.680 & 0.294 &\\
    \midrule
    \multirow{3}{*}{3} &$f_1$ & 22.70 & 0.684 & 0.287 & 
& 22.64 & 0.683 & 0.289 &\\
    &$f_2$ & 22.60 & 0.682 & 0.291 & 20\% 
& 22.58 & 0.680 & 0.294 & 37\% \\
     &$f_3$ & 22.64 & 0.680 & 0.295 &
& 22.47 & 0.675 & 0.303 &\\
     \midrule
    \multirow{3}{*}{4} &$f_1$ & 22.54 & 0.677 & 0.299 & 
& 22.52 & 0.678 & 0.298 &\\
    &$f_2$ & 22.28 & 0.670 & 0.305 & - 
& 22.31 & 0.672 & 0.304 & 15\%\\
     &$f_3$ & 22.62 & 0.673 & 0.308 & 
& 22.41 & 0.671 & 0.310\\
     \midrule
    \multirow{3}{*}{5} &$f_1$ & &  &  & 
& 22.45 & 0.673 & 0.305 &\\
    &$f_2$ &  &  &  & 
& 22.18 & 0.665 & 0.317 & -\\
     &$f_3$ &  &  &  & 
& 22.55 & 0.673 & 0.312 &\\\midrule
    \multicolumn{2}{c}{$\text{nerfacto}^\dagger$} & \tone 24.18 & \tone 0.729 & \tone 0.240 & - &
\tone 24.25 & \tone 0.732 & \tone 0.234 & -\\\bottomrule
    \end{tabular}
  \caption{\textbf{Detailed Results} for our capacity ablation study with the Mip-NeRF 360 dataset~\cite{Barron2022MipNeRF360}.
  In every configuration, our method performs best at $\ell = 1, f_2$.
  We color code {\colorbox{blue!42}{best}}, {\colorbox{blue!28}{second-best}} and {\colorbox{blue!16}{third-best}} for each scene.
  }
  \label{tab:supp:capacity-ablation}
\end{table}
\subsection{Impact of Capacity}
We provide the full results for our capacity experiment from the main material in Tab.~\ref{tab:supp:capacity-ablation}.
As we can see, layer $\ell = 1, f_2$ is the best configuration for our method across various scenarios.
This further indicates that capacity has minimal impact on our approach.

\subsection{ReLU or no ReLU}
We provide two example renderings comparing our method with the variant without the ReLU in Fig.~\ref{fig:relunorelu}.
As can clearly be seen, not using the ReLU leads to more artefacts, particularly around sharp edges, causing the drop in all image metrics.
This is caused by the more inconsistent histograms and the larger standard deviation.

\section{Per-Scene Results}
\label{sec:supp:per-scene}
In Tabs.~\ref{tab:supp:results-llff} and~\ref{tab:supp:results-blender}, we show per-scene results for the Blender dataset~\cite{MildenhallNeRF} and the LLFF dataset~\cite{mildenhall2019llff}. 
For the \emph{drums} scene, we manage to outperform Mip-NeRF quantitatively.
In Tabs.~\ref{tab:supp:results-mip360}~and~\ref{tab:supp:deep}, we also report the per-scene results for Mip-NeRF 360~\cite{Barron2022MipNeRF360} and Deep Blending~\cite{DeepBlending2018}.

\begin{table}[ht!]
  \scriptsize
  \centering
  \setlength{\tabcolsep}{3pt}
    \begin{tabular}{lccccccccc}\toprule
    \multicolumn{10}{c}{{$\textbf{NeRF}$}} \\\midrule
    $\ell$ & Fct. & Chair & Drums & Ficus & Hotdog & Lego & Materials & Mic & Ship\\\midrule
        \multirow{3}{*}{1} & $f_1$&30.10  &22.33  &26.21  &32.41  &26.79  &28.05  &29.10  &24.86  \\
     & $f_2$ &30.12  &23.30  &28.25  &33.40  &28.99  &28.58  &30.53  &26.30  \\
     & $f_3$ &30.43  &23.23  &28.14  &33.26  &28.75  &28.61  &30.48  &26.16  \\\midrule
    \multirow{3}{*}{2} & $f_1$ & \ttri 30.66  &23.57  &25.83  &33.17  &28.54  & \ttri 28.87  & \ttri 31.27  & \fail 22.04  \\
     & $f_2$ &30.19  & \ttri 23.63  & \ttri 28.45  & \ttri 33.61  & \ttri 30.03  &28.73  &30.97  & \ttri 26.79  \\
     & $f_3$ & \ttwo 30.69  & \ttwo 23.91  & \ttwo 29.03  & \ttwo 33.86  & \ttwo 30.62  & \ttwo 29.01  & \ttwo 31.37  & \ttwo 27.03  \\\midrule
    \multirow{3}{*}{3} & $f_1$ &27.93  &20.26  &21.74  & \fail 28.10 &\fail 20.56  &26.13  &28.15  & \fail  20.48  \\
     & $f_2$ &28.98  &20.46  &22.42  &32.97  &27.62  &26.18  &25.28  &25.91  \\
     & $f_3$ &27.93  &19.62  &21.29  &32.15  &26.14  &24.17  &23.59  &25.71  \\
    \midrule
        \multicolumn{2}{c}{$\text{NeRF}$} & \tone 31.16  & \tone 24.09  & \tone 29.26  & \tone 34.33  & \tone 31.34  & \tone 29.08  & \tone 31.75  & \tone 27.92  \\
    \midrule
    \multicolumn{10}{c}{{$\textbf{Mip-NeRF}$}} \\\midrule
        \multirow{3}{*}{1} & $f_1$ &30.14  &22.33  &26.10  &33.06 &27.52  &28.21  &28.54  &24.44  \\
     & $f_2$ & 30.01  &23.36  &27.60  &33.64  &29.83  &28.60  &30.25  &26.69  \\                 
     & $f_3$ &30.34  &23.34  &27.44  & \ttwo 33.87  &29.70  &28.61  &29.93  &26.56  \\\midrule
    \multirow{3}{*}{2} & $f_1$ & \ttwo 30.77  & \ttwo 23.91  &27.46  &31.67  &28.97  &27.97  & \ttri 30.99  & \fail 11.65  \\ 
     & $f_2$ &30.13  &23.66  & \ttri 28.07  &33.57  & \ttri 30.38  & \ttri 28.62  &30.97  &26.89  \\   
     & $f_3$ & \ttri 30.65  & \tone 23.95  & \ttwo 28.39  & \ttri 33.72  & \ttwo 30.82  & \ttwo 28.83  & \ttwo 31.30  & \ttwo 27.16  \\\midrule
    \multirow{3}{*}{3} & $f_1$ &28.77  & 22.31 & 25.96  & \fail 17.52  &\fail 23.23  &27.85  &30.90  &\fail 11.20  \\ 
     & $f_2$ &29.19  &22.68  &25.16  &33.60  &29.64  &27.75  &25.06  &26.66  \\                    
     & $f_3$ &28.07  &21.86  &23.22  &33.44  &29.64  &26.27  &21.98  & \ttri 27.10  \\                   
    \midrule
        \multicolumn{2}{c}{$\text{Mip-NeRF}$} & \tone 31.13  & \ttri 23.82  & \tone 28.52  & \tone 34.41  & \tone 31.07  & \tone 29.31  & \tone 31.54  & \tone 27.71  \\
    \midrule
    \multicolumn{10}{c}{{$\textbf{nerfacto}^\dagger$}} \\\midrule
    \multirow{3}{*}{1} & $f_1$  & 28.30  & 20.80 & 22.06  & 31.48  & 25.89  & 24.19  & 27.49  & 25.14  \\
     & $f_2$                    & 29.07  & 21.17  & \ttwo 22.36  & \ttri 31.98  &26.89  &24.35  &28.19  &25.56  \\
     & $f_3$                    & 29.07 & 21.18 & \ttri 22.21  & \ttwo 32.03  &26.91  &24.30  &28.02  &25.66  \\\midrule
    \multirow{3}{*}{2} & $f_1$  & \ttri 29.16  & 20.89 & 22.08  & 31.19  & 27.26  & 24.61  & 29.18  & 25.57  \\
     & $f_2$                    & \ttwo 29.35  & \ttwo 21.36  &22.17  &31.46  & \ttri 28.16  &24.56  &29.45  & \ttwo 25.93  \\
     & $f_3$                    & 28.56  & \ttri 21.24  &22.04  &31.18  &27.40  & \ttwo 24.66  & \ttri 29.51  & \ttri 25.92  \\\midrule
    \multirow{3}{*}{3} & $f_1$  & 27.96  & 20.59  &22.02  &31.00  & \ttwo 28.50  & \ttri 24.64  & \ttwo 29.52  &25.04  \\
     & $f_2$                    & 28.01  & 20.64  & 22.00  & 31.03  & 25.24  & 24.36  & 28.16  & 25.16  \\
     & $f_3$                    & 27.95  & 20.60  & 22.02  & 31.01  & 22.69  & 23.53  & 25.55  & 24.97  \\
    \midrule
        \multicolumn{2}{c}{$\text{nerfacto}^\dagger$} & \tone 30.64  & \tone 22.11  & \tone 22.86  & \tone 32.67  & \tone 30.65  & \tone 24.76  & \tone 29.78  & \tone 27.38  \\
        \bottomrule
    \end{tabular}
  \caption{\textbf{Per-Scene PSNR} results for the LLFF dataset~\cite{mildenhall2019llff}. 
  We color code {\colorbox{blue!42}{best}}, {\colorbox{blue!28}{second-best}} and {\colorbox{blue!16}{third-best}} for each scene. In addition, we highlight {\colorbox{red!42}{failures}} of our method.}
  \label{tab:supp:results-llff}
\end{table}

\begin{table}[t!]
  \scriptsize
  \centering
    \setlength{\tabcolsep}{3pt}
    \begin{tabular}{lccccccccc}\toprule
    \multicolumn{10}{c}{{$\textbf{NeRF}$}} \\\midrule
    $\ell$ & Fct. & Fern & Flower & Fortress & Horns & Leaves & Orchids & Room & T-Rex\\\midrule
        \multirow{3}{*}{1} & $f_1$ &23.83  &27.61  &28.12  &25.11  & \ttri 21.38  &19.35  &30.32  &25.16  \\
     & $f_2$ & \ttwo 24.19  & \ttwo 28.03  & \ttwo 28.27  & \ttwo 25.52  &21.35  &19.57  & \ttwo 30.81  & \ttwo 25.44  \\
     & $f_3$ & \ttri 24.07  & \ttri 27.97  & \ttri 28.22  & \ttri 25.42  & \ttwo 21.38  &19.48  & \ttri 30.73  & \ttri 25.43  \\\midrule
    \multirow{3}{*}{2} & $f_1$ &23.76  &27.68  &27.95  &24.84  &21.08  &19.48  &28.66  &24.32  \\
     & $f_2$ &24.03  &27.95  &28.17  &25.26  &21.23  & \ttri 19.62  &29.44  &25.30  \\
     & $f_3$ &23.92  &27.86  &28.09  &25.14  &21.19  &19.56  &29.13  &25.21  \\\midrule
    \multirow{3}{*}{3} & $f_1$ &23.61  &27.65  &27.74  &24.34  &20.97  &19.50  &28.18  &\fail 21.73  \\
     & $f_2$ &23.92  &27.95  &28.06  &24.96  &21.17  & \ttwo 19.65  &28.94  &25.26  \\
     & $f_3$ &23.82  &27.88  &27.95  &24.79  &21.12  &19.61  &28.66  &25.17  \\
    \midrule
        \multicolumn{2}{c}{$\text{NeRF}$} & \tone 24.70  & \tone 28.61  & \tone 29.07  & \tone 26.05  & \tone 21.56  & \tone 19.95  & \tone 31.49  & \tone 25.72  \\
    \midrule
    \multicolumn{10}{c}{{$\textbf{Mip-NeRF}$}} \\\midrule
    \multirow{3}{*}{1} & $f_1$ &23.56  &27.73  &28.04  &24.79  & \ttri 21.38  &19.34  &30.57  &25.16  \\
     & $f_2$ & \ttwo 23.93  &28.26  &28.50  & \ttri 25.51  &21.37  &19.59  &30.89  &25.56  \\             
     & $f_3$ &23.80  &28.16  &28.32  &25.35  & \ttwo 21.39  &19.50  &30.79  &25.46  \\\midrule
    \multirow{3}{*}{2} & $f_1$ & 23.41  &27.88  &28.13  &24.89  &21.23  &19.41  &30.47  &24.72  \\
     & $f_2$ &23.72  &28.33  &28.52  &25.49  &21.34  &19.60  & \ttwo 30.92  & \ttri 25.58  \\
     & $f_3$ &23.60  &28.29  &28.37  &25.35  &21.34  &19.53  &30.78  &25.51  \\\midrule
    \multirow{3}{*}{3} & $f_1$ &23.51  &25.69  &28.31  &23.68  &20.74  &18.75  &30.14  &\fail 15.30  \\
     & $f_2$ & \ttri 23.83  & \ttwo 28.37  & \ttwo 28.72  & \ttwo 25.52  &21.32  & \ttwo 19.67  & \ttri 30.90  & \ttwo 25.61  \\
     & $f_3$ &23.71  & \ttri 28.34  & \ttri 28.60  &25.37  &21.31  & \ttri 19.61  &30.81  &25.57  \\\midrule
        \multicolumn{2}{c}{$\text{Mip-NeRF}$} & \tone 24.54  & \tone 28.69  & \tone 29.47  & \tone 26.13  & \tone 21.59  & \tone 20.01  & \tone 31.59  & \tone 25.92  \\
    \midrule
    \multicolumn{10}{c}{{$\textbf{nerfacto}^\dagger$}} \\\midrule
     \multirow{3}{*}{1} & $f_1$  & \ttri 23.52  &26.60  & \ttri 27.59  &25.18  &17.98  & \ttri 19.03  &29.81  & \ttri 25.95  \\
     & $f_2$                  & \ttwo 23.72  & \ttwo 26.78  & \ttwo 27.70  & \ttri 26.44  &17.86  & \ttwo 19.06  &30.09  &25.88  \\
     & $f_3$                   &23.10  &26.30  &27.22  &24.88  &18.14  &18.92  &29.45  &25.06  \\\midrule
    \multirow{3}{*}{2} & $f_1$  &22.32  &26.06  &26.84  &25.97  & \ttri 18.61  &18.57  & \ttri 30.20  &25.07  \\
     & $f_2$                    &22.98  & \ttri 26.64  &27.53  & \ttwo 26.65  & \ttwo 19.25  &18.99  & \ttwo 30.28  & \ttwo 26.34  \\
     & $f_3$                    &21.97  &25.21  &26.45  &23.85  &18.29  &18.33  &29.98  &24.03  \\\midrule
    \multirow{3}{*}{3} & $f_1$  &22.44  &25.83  &27.23  &23.65  &18.34  &17.88  &28.03  &22.28  \\
     & $f_2$                    &22.06  &25.96  &26.76  &24.58  &18.46  &17.77  &28.81  &25.87  \\
     & $f_3$                    &22.13  &25.77  &27.05  &23.51  &18.32  &17.40  &27.62  &22.49  \\
    \midrule
        \multicolumn{2}{c}{$\text{nerfacto}^\dagger$} & \tone 23.83  & \tone 27.01 & \tone 27.94& \tone 26.91  & \tone 21.28  & \tone 19.11 & \tone 30.36  & \tone 26.39 \\
        \bottomrule
    \end{tabular}
  \caption{\textbf{Per-Scene PSNR} results for the Blender dataset~\cite{MildenhallNeRF}. 
  We color code {\colorbox{blue!42}{best}}, {\colorbox{blue!28}{second-best}} and {\colorbox{blue!16}{third-best}} for each scene. In addition, we highlight {\colorbox{red!42}{failures}} of our method.}
  \label{tab:supp:results-blender}
\end{table}

\begin{table}[t!]
  \footnotesize
  \centering
    \begin{tabular}{@{}lcccccc@{}}
\toprule
&& \multicolumn{5}{c}{Outdoor Scenes} \\ \cmidrule{3-7}
$\ell$ & Fct. & Bicycle & Flowers & Garden & Stump & Treehill \\
\midrule
\multirow{3}{*}{1} & $f_1$ & 20.47  & \ttri 20.15 & 23.69 &  22.88 & \ttri 17.55 \\
& $f_2$ & \ttwo 20.70  & \ttwo 20.17 & \ttwo 23.75 &  \ttwo 22.98 & \ttwo 17.55 \\
& $f_3$  & 20.51  & 20.14 & \ttri 23.69  & 22.90 & 17.54 \\
\midrule
\multirow{3}{*}{2} & $f_1$  & 20.45  & 20.13 & 23.58  & 22.89 & 17.35 \\
& $f_2$ & \ttri 20.59 & 20.03 & 23.45 & \ttri 22.97 & 17.33 \\
& $f_3$ & 20.41 & 20.12 & 23.61  & 22.86 & 17.46 \\
\midrule
\multirow{3}{*}{3} & $f_1$ & 24.52 &  19.98 & 23.48  & 22.71 & 17.47 \\
& $f_2$ & 20.03  & 19.94 & 23.20  & 22.56 & 17.42 \\
& $f_3$ & 20.12 & 19.89 & 23.54  & 22.70 & 17.48 \\
\midrule
\multicolumn{2}{c}{$\ourfacto$} & \tone 22.19  & \tone 20.81 & \tone 24.87 & \tone 24.03 & \tone 17.88 \\
\midrule
&&& \multicolumn{4}{c}{Indoor Scenes} \\ \cmidrule{4-7}
&&& Bonsai & Counter & Kitchen & Room \\\midrule
\multirow{3}{*}{1} & $f_1$ && 25.43 & \ttri 24.06  & 24.29 & 26.66  \\
& $f_2$  && \ttwo 25.65 & \ttwo 24.20 & \ttwo 24.54 & \ttwo 26.79  \\
& $f_3$   && 25.40 & 24.00 & \ttri 24.32 & \ttri 26.73  \\
\midrule
\multirow{3}{*}{2} & $f_1$ && \ttri 25.47 & 23.96  & 23.97 & 26.56 \\
& $f_2$ && 25.41 & 23.99 & 23.68 & 26.47  \\
& $f_3$ && 25.15 & 23.94 & 23.93 & 26.52  \\
\midrule
\multirow{3}{*}{3} & $f_1$ && 24.52 & 23.76 & 23.82 & 26.03  \\
& $f_2$ && 24.36 & 23.64 & 23.15 & 25.63  \\
& $f_3$ && 24.29 & 23.90 & 23.63 & 25.96 \\
\midrule
\multicolumn{2}{c}{$\ourfacto$} && \tone 27.55 & \tone 25.36 & \tone 26.76 & \tone 28.37 \\\bottomrule
\end{tabular}
  \caption{\textbf{Per-Scene PSNR} results for the Mip-NeRF 360 dataset~\cite{Barron2022MipNeRF360}. 
  We color code {\colorbox{blue!42}{best}}, {\colorbox{blue!28}{second-best}} and {\colorbox{blue!16}{third-best}} for each scene.}
  \label{tab:supp:results-mip360}
\end{table}

\begin{table}[ht!]
  \footnotesize
  \setlength{\tabcolsep}{5pt}
  \centering
    \begin{tabular}{@{}llrrrrrr@{}}
    \toprule
         && \multicolumn{3}{c}{Dr Johnson} & \multicolumn{3}{c}{Playroom} \\
         \cmidrule(lr){3-5}\cmidrule(lr){6-8}
     && PSNR & SSIM & LPIPS  & PSNR & SSIM & LPIPS \\
    \midrule
\multirow{3}{*}{1} &$f_1$ & 28.39 & 0.885 & 0.195 & \ttri 27.63 & \ttri 0.831 & \ttri 0.280\\
&$f_2$ & 28.39 & 0.885 & 0.196 & \ttwo 27.64 & \ttwo 0.833 & \ttwo 0.279 \\
&$f_3$ & 28.42 & 0.885 & \ttri 0.194 & 27.55 & 0.830 & 0.284  \\
\midrule
\multirow{3}{*}{2} &$f_1$ & \ttri 28.46 & \ttwo 0.886 & \ttwo 0.194 & 27.49 & 0.830 & 0.284  \\
&$f_2$ & \ttwo 28.47 & \ttri 0.886 & 0.194  & 27.44 & 0.827 & 0.291  \\
&$f_3$ & 28.38 & 0.885 & 0.196  & 27.49 & 0.828 & 0.289 \\
\midrule
\multirow{3}{*}{3} &$f_1$ & 28.23 & 0.881 & 0.202  & 27.30 & 0.818 & 0.310  \\
&$f_2$ & 28.03 & 0.879 & 0.205 & 26.93 & 0.809 & 0.322  \\
&$f_3$ & 28.18 & 0.877 & 0.209  & 27.38 & 0.817 & 0.315 \\
\midrule
\multicolumn{2}{c}{$\ourfacto$} & \tone 29.65 & \tone 0.903 & \tone 0.171 & \tone 28.87 & \tone 0.856 & \tone 0.242 \\
    \bottomrule
    \end{tabular}
  \caption{\textbf{Full Per-Scene Results} for the Deep Blending dataset~\cite{DeepBlending2018}.
  We color code {\colorbox{blue!42}{best}}, {\colorbox{blue!28}{second-best}} and {\colorbox{blue!16}{third-best}} for each scene.
  }
  \label{tab:supp:deep}
\end{table}

\end{document}